\documentclass{article}
\usepackage{siunitx}
\usepackage{booktabs}
\usepackage{mathtools} 
\usepackage{bbm}
\usepackage{amsmath}
\usepackage[utf8]{inputenc}
\usepackage{graphicx}           
\usepackage{geometry}           
\usepackage{authblk}            
\usepackage{amsmath}            
\usepackage{amssymb}            
\usepackage{bbm}                
\usepackage{graphicx}           
\usepackage{multirow}           
\usepackage{mhchem}             
\usepackage{tikz}               
\usepackage[authoryear,round]{natbib}

\usepackage[ruled,vlined,linesnumbered]{algorithm2e}
\usepackage{hyperref}
\DeclareUnicodeCharacter{202F}{\,}

\title{HOFLON: Hybrid Offline Learning and Online Optimization for Process Start‑Up and Grade‑Transition Control}
\author[1]{Alex Durkin}
\author[2]{Jasper Stolte}
\author[1]{Mehmet Mercang\"{o}z}
\affil[1]{Department of Chemical Engineering, Imperial College London, SW7 2AZ, UK}
\affil[2]{Shell Information Technology International BV, 1031 HW Amsterdam, NL}
\date{}

\begin{document}
 
\maketitle
 
\section*{Abstract}
 
Start‑ups and product grade‑changes are critical steps in continuous‑process plant operation, because any misstep immediately affects product quality and drives operational losses. These transitions have long relied on manual operation by a handful of expert operators, but the progressive retirement of that workforce is leaving plant owners without the tacit know‑how needed to execute them consistently. In the absence of a process model, offline reinforcement learning (RL) promises to capture—and even surpass—human expertise by mining historical start‑up and grade‑change logs, yet standard offline RL struggles with distribution‑shift and value‑overestimation whenever a learned policy ventures outside the data envelope. We introduce HOFLON (Hybrid Offline Learning + Online Optimization) to overcome those limitations. Offline, HOFLON learns (i) a latent data manifold that represents the feasible region spanned by past transitions and (ii) a long‑horizon Q‑critic that predicts the cumulative reward from state–action pairs. Online, it solves a one‑step optimization problem that maximizes the Q‑critic while penalizing deviations from the learned manifold and excessive rates of change in the manipulated variables. We test HOFLON on two industrial case studies—a polymerization reactor start‑up and a paper‑machine grade‑change problem—and benchmark it against Implicit Q‑Learning (IQL), a leading offline‑RL algorithm. In both plants HOFLON not only surpasses IQL but also delivers on average better cumulative rewards compared to the best start‑up or grade‑change ever observed in the historical data, demonstrating its potential to automate transition operations beyond current expert capability.

 
\section{Introduction}
 
Start-up and product grade-change operations are critical transient phases in continuous process plants. These differ fundamentally from maintaining a stable steady-state by exhibiting a definable initial condition and a conclusive final state, thereby requiring distinct dynamic control strategies \citep{kruger2004optimization}. Any misstep during these transients immediately affects product quality, causes off-spec material, and leads to economic losses. For example, polymer reactors experience considerable off-spec polymer during transitions due to nonlinear reactor behavior \citep{kao2025control}. Achieving rapid, safe, and constraint-compliant transitions provides a significant competitive advantage. Historically, these operations relied on tacit knowledge of experienced operators. However, many senior operators are retiring, creating a significant knowledge gap. This tacit knowledge, difficult to document and standardize, results in operational variability and potential safety risks. Consequently, there is a pressing need to codify this expertise systematically to help junior process operators or to enable autonomous operations.

Procedural automation and state-based control have been widely adopted to ensure consistent execution of start-up operations in steps but these procedures cannot capture complex control trajectories \citep{chang2021process}. Advanced control approaches such as Model Predictive Control (MPC) and dynamic optimization have also been employed to optimize grade transitions in polymer and paper industries. However, these approaches require accurate mechanistic models, which are challenging and resource-intensive to develop and might not be always feasible to deploy \citep{Chu2019}.

Recent trends emphasize leveraging historical operational data through statistical and data-driven techniques. Methods like multi-way principal component analysis (PCA) and partial least squares (PLS) effectively identify optimal operational patterns from historical logs without the need for detailed mechanistic models \citep{damodaran2015monitoring}. Data-driven soft sensors predicting real-time product quality during transitions are also increasingly adopted \citep{choi2021inferential}. However, such statistical methods generally struggle to represent complex, dynamic relationships between process measurements and manipulated variables, especially when precise control along specific trajectories is required to achieve optimal transitions.

\subsection{Control of polymerization reactors}

Polymer reactor start‑ups and grade switches are inherently nonlinear, multivariable operations; if executed slowly or inaccurately they generate large volumes of off‑spec product and erode profit \citep{bonvin2005optimal,lee2003iterative}. Because each minute of transition time produces waste, operators sometimes overshoot or undershoot set‑points to accelerate the change, risking constraint violations and safety incidents \citep{prata2008integrated}. Hence grade transitions impose a dual burden: complex dynamics with tight constraints, and the economic drive to minimise downtime and scrap. Steady‑state quality alone is insufficient---manufacturers must also perform grade changes quickly and reliably to meet market demands \citep{bonvin2005optimal}.

Early solutions relied on first‑principles modeling and optimal‑control formulations that computed feed, hydrogen, and temperature trajectories to minimize time or off‑spec volume \citep{prata2008integrated}. Model predictive control (MPC) soon became the implementation vehicle, coordinating multivariable actions within safety limits. Full‑scale MPC/APC deployments in polyethylene plants, for example, report 25-50\% faster transitions, lower scrap, and roughly 7\% higher throughput \citep{bonvin2005optimal}.

More recently, data‑centric methods augment or replace physics‑based approaches. Iterative learning control (ILC) refines repeated grade changes, shrinking error and waste run by run \citep{bristow2006survey,lee2003iterative}. Machine‑learning models---especially hybrid neural‑network ``digital twins'' that blend mechanistic balances with data‑driven kinetics---feed into nonlinear MPC and provide virtual sensors for on‑the‑fly quality prediction \citep{daoutidis2024machine,chang2022grade,benamor2004polymer}. These adaptive strategies push grade‑change automation toward shorter, safer, and more economical transitions while reducing reliance on perfect first‑principles knowledge.

\subsection{Control of paper machine grade transitions}

Transitions from one paper grade to another are among the most disruptive events in papermaking: they create off‑spec production, downtime, and quality variability. Early work focused on dynamic models of the machine during transitions \citep{Chen1995}, enabling the first optimization‑based actuator and set‑point trajectories \citep{Ihalainen1996}. With the rise of multivariable predictive control, industrial case studies showed that coordinated, model‑based transition control can shorten change times by up to 30\% and cut variability in basis weight and moisture \citep{Murphy1999}. Recent advances go further: data‑driven schedulers minimize \emph{how often} grade changes occur by optimizing production sequences \citep{Mostafaei2020}, while nonlinear MPC packages inside modern quality‑control systems provide operator‑friendly one‑button transitions on the machine floor. Field reports from large mills confirm that combining well‑maintained grade‑change libraries with continuous monitoring sustains 28-35\% faster transitions over the long term \citep{Chu2019}. Despite these automation gains, experienced operators remain pivotal: they validate soft‑sensor predictions, intervene during upsets, fine‑tune set‑points between grades, and supply the knowledge that shapes and updates transition libraries\citep{Vipetec2025}.
 
\subsection{Offline reinforcement learning}
Reinforcement learning is the field within machine learning in which an agent learns to optimize its behavior in an environment through interaction \citep{sutton2018reinforcement}. The key advantage of reinforcement learning is that it can learn a control policy which deals with the trade off between greediness and patience to chase an even greater expected reward in the future. The theoretical framework underpinning reinforcement learning is the Markov Decision Process (MDP). From a process systems perspective, one major challenge for RL in process plants is the development of realistic and representative simulation environments that capture the nonlinear, multivariable nature of chemical processes, including reaction kinetics, heat and mass transfer, and operational constraints. Safety and stability requirements further complicate the learning process, as policies must not only optimize performance but also ensure that operating conditions remain within safe and feasible bounds \citep{baldea2025automated}.

In offline reinforcement learning \citep{levine2020offline, prudencio2023survey} a buffer of historic observations is used to train a policy for the agent. In this setting the data was collected under an unknown behavior policy $\pi_\beta$, which could be any combination of manual operation with supervisory control. The objective is to learn a safe and robust control policy from this data without further environment interaction.
 
The key challenge with offline reinforcement learning is \emph{distributional shift} \citep{kumar2019stabilizing}, which can be conceptually understood as the difficulty to accurately estimate the value of state/action combinations that were not observed in the historic data. To illustrate, consider the temporal difference loss to minimize the Bellman error for the Q-function as given below:
 
\begin{equation} \label{eq:bellman}
    \mathcal{L}(\theta) = \mathbb{E}_{\left(s, a, s'\right) \sim \mathcal{D}} \left[ \left( r(s,a) + \gamma \max_{a'} Q_{\hat{\theta}}(s', a') - Q_{\theta}(s, a) \right)^2 \right]
\end{equation}
 
where $\mathcal{D}$ is the dataset of (state, action, next state) observations, $Q_{\theta}$ is the parameterized Q-function prediction for the current state-action pair, and $r(s,a) + \gamma \max_{a'} Q_{\hat{\theta}}$ is the Bellman equation for calculating the target Q-value as the sum of the immediate reward and the discounted maximum expected future rewards.
 
The challenge with distributional shift is that the loss in equation \ref{eq:bellman} contains a term $Q_{\hat{\theta}}(s',a')$ which requires evaluating the Q-function at state/action combinations that are not present in the data. Even worse, this Q-function evaluation is maximized over all possible future actions leading to gross value overestimation, learning policies that exploit this non-existing value.
 
Typical strategies to stabilize learning as employed in online reinforcement learning include double Q-learning \citep{vanhasselt2015deep} and using target networks \citep{mnih2015human} or simply using vast amounts of training data.
These strategies can only slightly help reduce the magnitude of the value overestimation but do not fundamentally solve the problem.
Since this issue was clearly articulated in \citep{kumar2019stabilizing} in 2019 the field of offline reinforcement learning gathered momentum and many strategies have been devised to develop policies with more robustness towards distributional shift.
Algorithms such as Conservative Q-Learning (CQL) \citep{kumar2020conservative}, Implicit Q-Learning (IQL) \citep{kostrikov2021offline}, and Behavior-Regularized Actor-Critic (BRAC) \citep{wu2019behavior} mitigate distributional shift by constraining policy updates toward the behavior policy or learning conservative value estimates.
 
\subsubsection{Implicit Q-learning}
Implicit Q-learning (IQL) \citep{kostrikov2021offline} is an offline RL approach that avoids the distributional shift problem by never querying the target Q-function for state/action combinations that are not in the dataset. This fully in-sample learning depends on an approximation of the Q-value of the next state/action combination by a state-value function. Expectile regression is used to predict an upper expectile of the state-value function with respect to the action distribution:
 
\begin{equation}
    \mathcal{L}_V(\psi) = \mathbb{E}_{\left(s, a\right) \sim \mathcal{D}} \left[ \mathcal{L}_2^\tau \left( Q_{\hat{\theta}}(s, a) - V_{\psi}(s) \right) \right]
\end{equation}
 
where $\mathcal{L}_2^\tau(u) = |\tau - \mathbbm{1}(u<0)|u^2$ is the expectile loss function, which is a generalization of the mean squared error loss which enables asymmetric weightings to favour larger expectiles. If $\tau$ is equal to $0.5$, the expectile regression is identical to the normal MSE. As $\tau$ is increased towards $1$ the value no longer learns the average value of a given state over the action distribution but approaches the maximum value supported by the actions in the data. This estimated value function is then used to backup into the Q-function update instead of $Q_{\hat{\theta}}(s',a')$:
 
\begin{equation}
    \mathcal{L}_Q(\theta) = \mathbb{E}_{\left(s, a, s'\right) \sim \mathcal{D}} \left[ \left( r(s,a) + \gamma V_{\psi}(s') - Q_{\theta}(s, a) \right)^2 \right]
\end{equation}
 
Once an estimated Q-value function is learned, the policy maximizing the expected value is extracted using advantage weighted regression (AWR) \citep{peng2019advantage}, which also only evaluates the Q-function at state/action combinations that are present in the dataset:
 
\begin{equation}
    \mathcal{L}_\pi(\phi) = \mathbb{E}_{(s, a) \sim \mathcal{D}} \left[ \exp(\beta(Q_{\hat{\theta}}(s, a) - V_\psi(s)))\log\pi_\phi(a | s) \right]
\end{equation}
 
where $\beta$ is an inverse temperature parameter which controls the trade-off between maximizing the Q-function ($\beta\rightarrow\infty$) and staying close to the behavior policy ($\beta\rightarrow0$) thereby supporting a spectrum of risk sensitivities.
AWR is based on a forward KL divergence regularization pulling the policy towards actions observed in the data, pulling more strongly towards actions of higher estimated value.

\subsubsection{Other approaches for handling out-of-distribution actions in offline RL}

Early generative modeling approaches, particularly autoencoders, have been widely utilized to constrain policy outputs within the training dataset's action distribution. For instance, Batch-Constrained Q-learning (BCQ) \citep{fujimoto2019off} employs a conditional variational autoencoder (VAE) trained on the offline dataset, restricting action selection to within dataset support and preventing unrealistic extrapolations. Similarly, Policy in Latent Action Space (PLAS) \citep{zhou2020plas} leverages a latent action representation learned via a VAE, ensuring that generated actions remain inherently bounded within the support of available data. While these autoencoder-based methods effectively mitigate OOD queries, their conservative nature can also limit beneficial policy improvements due to overly strict constraints.

Recently, diffusion models have emerged as powerful alternatives capable of capturing complex, multimodal action distributions. Diffusion-based methods, such as Diffusion-QL \citep{wang2023diffusion}, learn action distributions through iterative denoising processes, inherently generating actions that closely adhere to the dataset distribution. However, these methods initially exhibited high computational overhead. Subsequent advancements like Efficient Diffusion Policies (EDP) \citep{kang2023edp} significantly reduced training complexity by reconstructing actions from partially denoised states. Trajectory-level approaches, exemplified by Diffuser \citep{janner2022diffuser}, extend this concept by planning entire action sequences conditioned on desired outcomes, thus inherently curbing the occurrence of OOD actions.

Current research directions further enhance diffusion models by integrating value-aware guidance mechanisms. Advantage-Weighted Diffusion Actor-Critic (ADAC) \citep{chen2025adac} combines diffusion-based action generation with advantage-weighted value updates, selectively promoting potentially advantageous OOD actions while penalizing harmful ones. Diffusion-DICE \citep{mao2024diffusiondice} introduces a two-stage guidance process, first generating candidate actions within the dataset distribution and subsequently selecting optimal candidates based on critic evaluations, thereby effectively minimizing OOD errors. Additionally, prior-guided diffusion planning methods \citep{ki2025prior} replace standard Gaussian priors with learned priors concentrating diffusion processes on high-value, in-distribution trajectories. Collectively, these methods illustrate a principled evolution toward robust offline RL by leveraging generative modeling frameworks to balance action conservatism with the capacity for meaningful policy improvement.

\subsubsection{Q\texorpdfstring{\,-}{ }Learning and direct optimization of the action--value function for continuous control} 
A central challenge in applying reinforcement learning to problems with continuous action spaces is the selection of the optimal action. Given a learned action-value (or Q-) function, $Q(s, a)$, which estimates the long-term return from taking action $a$ in state $s$, the agent must find the action that maximizes this function at each decision step. The dominant paradigm for addressing this challenge is the \textbf{actor-critic framework} \citep{lillicrap2016ddpg}. In this approach, a separate "actor" network, $\pi_\phi(s)$, is trained to approximate the $\arg\max_a Q_\theta(s,a)$ operation. At runtime, action selection is a fast, single forward pass through the actor network. While computationally efficient, this introduces a second, interdependent learning problem that is notoriously difficult to stabilize. More critically for offline learning, the actor can become detached from the critic, learning to exploit regions where the Q-function erroneously predicts high values due to extrapolation error beyond the support of the offline dataset. An alternative and increasingly powerful paradigm is to forgo the actor network entirely and instead perform \textbf{direct optimization of the Q-function} at runtime. In this "critic-only" approach, the learned $Q_\theta(s,a)$ is treated as an objective function, and a numerical solver is used to find the maximizing action at each step: \[ a_k^\star = \arg\max_{a \in \mathcal{A}} Q_\theta(s_k, a). \] This approach offers several compelling advantages, particularly for offline and safety-critical settings. First, it eliminates the instabilities associated with training a separate actor. Second, it provides a natural mechanism for enforcing constraints on the action space. Third, and most relevant to our work, it allows for the inclusion of additional objectives and regularizers directly into the action-selection process, enabling fine-grained control over the policy's behavior. The primary challenge of this approach is the non-convex nature of the maximization problem itself. A rich body of recent work has focused on developing effective methods to solve or approximate this optimization. \paragraph{Tackling the Maximization Challenge.} Research into direct Q-function optimization has explored several distinct strategies. One direction seeks to make the optimization tractable by design. \textbf{Structured parameterizations} impose constraints on the Q-function's architecture; for example, Normalized Advantage Functions (NAF) use a quadratic form for the advantage to yield an analytical solution \citep{gu2016naf}, while Input-Convex Neural Networks ensure that any local optimum found is global \citep{amos2017icnn}. When the Q-function is an unconstrained deep neural network, the maximization can be addressed with \textbf{stochastic search and sampling methods}. \textsc{QT-Opt} successfully used a cross-entropy method (CEM) for robotic grasping without any actor network \citep{kalashnikov2018qtopt}. Soft Q-Learning (SQL) reinterprets the Q-function as an energy landscape and samples high-value actions from the corresponding Boltzmann distribution, avoiding the hard maximization entirely \citep{haarnoja2017sql}. For applications where optimality is critical, \textbf{exact mathematical programming} offers a powerful solution. If the Q-function is a ReLU network, the maximization can be formulated as a mixed-integer linear program (MILP) and solved to global optimality, producing highly accurate targets and superior control performance \citep{ryu2020caql, yuan2023milpq}. These articles demonstrates that by framing action selection as an explicit optimization problem, we gain the ability to incorporate domain knowledge and safety considerations for the solution. This paradigm provides an important element for the approach we present in this paper. 

\subsection{The HOFLON RL approach}

Our approach is designed for a common yet challenging scenario in industrial control: we lack a first-principles dynamic model or a high-fidelity simulator, but have access to a rich historical dataset of past operations. Critical procedures like process start-ups and product grade-changes often fall into this category, where logs capture the operational history, whether generated by experienced human operators or existing control systems. The core opportunity lies in leveraging this data not merely to imitate past behavior, but to synthesize a novel control strategy that exceeds the performance of even the best historical run by intelligently recombining learned sub-policies.

To achieve this, a controller must explicitly balance three competing objectives: maximizing long-term performance, ensuring actions remain within a safe operating envelope defined by the data, and maintaining smooth actuator movements to prevent equipment wear. Standard offline RL algorithms attempt to solve this implicitly by regularizing a single, monolithic policy network, which often leads to opaque failure modes and makes it difficult to independently tune these critical trade-offs.

\begin{figure}[h]
    \centering
    \includegraphics[width=1\textwidth]{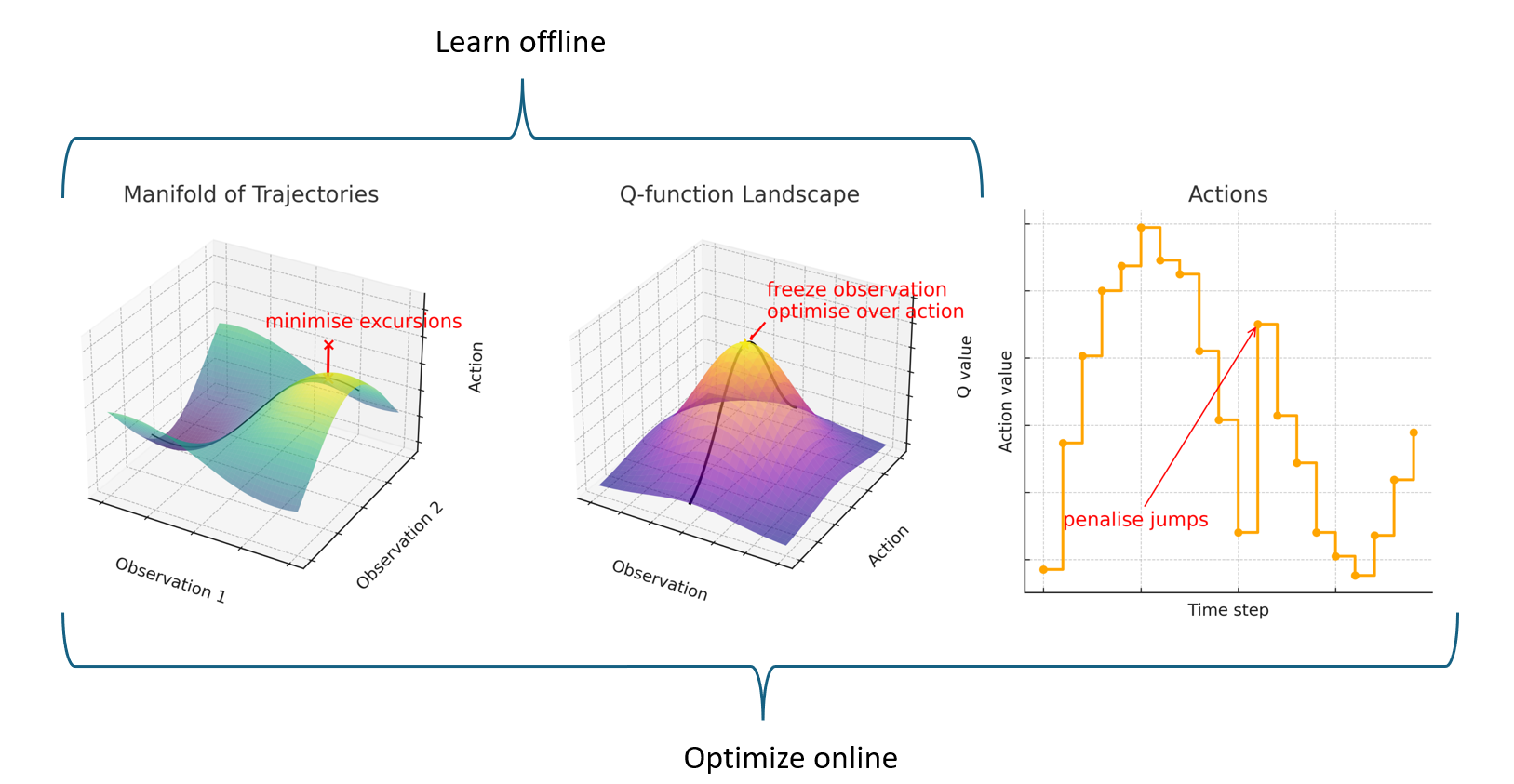}
    \caption{Conceptual representation of the elements in the HOFLON-RL algorithm.}
    \label{fig:HOFLON Comic}
\end{figure}

To address this challenge, we designed HOFLON based on a principle of explicit decomposition. Instead of learning a single policy, our approach decouples the problem into distinct, interpretable components. Crucially, we learn both the Q-critic (what actions lead to high value) and the data manifold model (what actions are safe) offline as functions of both state and action, capturing their complex joint relationship. During online deployment, the current state is observed and treated as a fixed parameter, transforming the learned multi-variable functions into well-defined objectives over the action space alone. These objectives are then combined with a third component—a direct regularizer on the rate of change of actions—within a real-time optimization framework. This explicit structure provides a transparent and robust framework for reasoning about performance, safety, and smoothness as independent, tunable goals, enabling the controller to discover and execute superior control sequences. A graphical representation of the HOFLON approach to offline RL is depicted in Fig. \ref{fig:HOFLON Comic}.

\subsection{Contributions}
This paper makes the following primary contributions to the field of data-driven process control and offline reinforcement learning: \begin{enumerate} \item \textbf{A Novel Hybrid Control Framework (HOFLON):} We introduce a new framework that synergizes the strengths of offline reinforcement learning and online, solver-based optimization. By decoupling the learning of a long-horizon value function from the real-time action selection, HOFLON avoids the instabilities of traditional actor-critic methods and provides a robust structure for deploying learned policies in safety-critical environments. \item \textbf{Latent Manifold Regularization for Safe Policy Execution:} We propose a specific mechanism to combat the critical challenge of distributional shift in offline RL. An adversarially trained autoencoder learns a model of the safe operating manifold from historical data. This model is then used online as a soft constraint, or regularizer, within the optimization problem, effectively preventing the controller from taking unsafe or unexplored actions. \item \textbf{State-of-the-Art Performance on Industrial Benchmarks:} We provide extensive validation of HOFLON on two high-fidelity industrial simulations: a polymerization reactor start-up and a paper machine grade-change. Our results demonstrate that HOFLON not only significantly outperforms a leading offline RL baseline (IQL) but also achieves performance superior to the best-ever recorded transition in the historical logs, showcasing its capability to automate complex operations beyond existing expert levels. \end{enumerate} The remainder of this paper is organized as follows. Section~\ref{sec:algorithm} details the HOFLON algorithm. Section~\ref{sec:dev_cstr} describes the process benchmarks. Section~\ref{sec:results} presents our experimental results, followed by a discussion in Section~\ref{sec:Discussion} . Finally, Section~\ref{sec:Conclusions} concludes the paper. 
 
\section{Algorithm development} \label{sec:algorithm}

We introduce a novel control algorithm, \textbf{H}ybrid \textbf{O}ffline \textbf{L}earning with \textbf{ON}line \textbf{R}einforcement \textbf{L}earning (\textbf{HOFLON-RL}). This method synthesizes a data-driven control policy by combining offline training on historical data with real-time, solver-based optimization. The architecture is designed for complex, continuous-action control problems where safety and performance are paramount, such as industrial process control. It avoids the instabilities of traditional actor-critic methods by directly optimizing a learned action-value function, regularized by a model of the safe operating envelope.

\subsection{Problem formulation}
\label{sec:problem_formulation}

We model the controller-plant interaction as a Markov Decision Process (MDP), defined by the tuple $\mathcal{M}=(\mathcal{S},\mathcal{A},p,r,\gamma)$, where $\mathcal{S}$ is the state space, $\mathcal{A}$ is the continuous action space, $p(s_{k+1}|s_k, a_k)$ is the unknown state transition dynamics, $r(s_k, a_k)$ is the reward function, and $\gamma \in (0, 1)$ is the discount factor.

At each discrete time step $k$, the controller observes the plant state $s_k \in \mathcal{S} \subset \mathbb{R}^d$ and selects an action $a_k \in \mathcal{A} \subset [0,1]^m$. The actions are scaled to physical units before being applied to the plant. The system then transitions to a new state $s_{k+1}$ and yields a reward $r_k$. The objective is to find a policy $\pi(a_k|s_k)$ that maximizes the expected discounted return $G_k = \mathbb{E}\left[\sum_{j=0}^{\infty} \gamma^j r_{k+j}\right]$, subject to two critical constraints:
\begin{enumerate}
    \item \textbf{Actuation Limits:} Actions must remain within hard physical bounds, $a_k \in [0,1]^m$.
    \item \textbf{Distributional Constraint:} The policy must only select actions that keep the system within the state-action distribution of the available historical data, ensuring safe and predictable behavior.
\end{enumerate}

\subsection{HOFLON-RL framework}
\label{sec:methodology}

The HOFLON-RL framework is divided into two distinct phases: an offline learning phase where historical data is used to train two key models, and an online optimization phase where these models are deployed to make real-time control decisions.

\subsubsection{Offline learning phase}
\label{sec:offline}

Given a static, offline dataset of historical transitions $\mathcal{D} = \{(s_k, a_k, r_k, s_{k+1})\}_{k=1}^{N}$, collected under one or more unknown behavioral policies, we learn two components: an action-value function (the Q-critic) that predicts long-term returns, and a generative model (the AAE) that captures the data distribution.

\paragraph{1. Learning the action-value function (Q-Critic).}
To guide the online optimization towards high-return actions, we train a Q-critic network, $\hat{Q}_{\theta}(s, a)$, to approximate the true action-value function $Q^*(s, a)$. Many offline RL algorithms learn the Q-function via Bellman bootstrapping. However, for finite-horizon episodic tasks common in grade transitions and startups, we can directly estimate the value function by regressing against the total empirical return accessible in historical data, which leads to stable solutions.

First, we compute the discounted return for each trajectory in the dataset:
\begin{equation}
    G_k = \sum_{j=0}^{H-1} \gamma^j r_{k+j},
    \label{eq:mc_returns}
\end{equation}
where $H$ is the episode horizon. The Q-critic $\hat{Q}_{\theta}$ is then trained via supervised learning to minimize the mean squared error:
\begin{equation}
    \mathcal{L}_Q(\theta) = \mathbb{E}_{(s_k, a_k, G_k) \sim \mathcal{D}} \left[ \left( G_k - \hat{Q}_{\theta}(s_k, a_k) \right)^2 \right].
    \label{eq:q_loss}
\end{equation}
The input to the Q-critic is an augmented state-action vector $z_k = [s_k, e_k, I_k, a_k]$, where $e_k = h(s_k, s_{\text{sp}})$ is the tracking error and $I_k = \sum_{i=0}^{k} e_i \Delta t$ is the integrated error, providing the critic with task-relevant context.

\paragraph{2. Learning the data manifold (Adversarial Autoencoder).}
To enforce the distributional constraint, we must be able to identify and penalize out-of-distribution (OOD) actions. We achieve this by training an Adversarial Autoencoder (AAE) on the state-action pairs from the historical data $\mathcal{D}$. The AAE consists of an encoder $E$ and a decoder $D$. The encoder maps an input state-action pair $(s, a)$ to a latent code $z = E(\sigma(s, a))$, where $\sigma(\cdot)$ is a feature-wise standardizer. The decoder reconstructs the original pair $\hat{a}, \hat{s} = D(z)$.

The AAE is trained with a composite loss function that combines a reconstruction objective with an adversarial regularizer, which forces the latent space to match a prior distribution (e.g., $\mathcal{N}(0,I)$):
\begin{equation}
    \mathcal{L}_{\text{AAE}} = \underbrace{\left\| [\sigma(s, a)] - D(E(\sigma(s, a))) \right\|_2^2}_{\text{Reconstruction Loss}} + \beta \cdot \underbrace{\text{JS}\left( q(z) \,\|\, p(z) \right)}_{\text{Adversarial Regularizer}}.
\end{equation}
Once trained, the AAE provides a powerful OOD regularizer. We define the \emph{latent inconsistency penalty} as the reconstruction error:
\begin{equation}
    \phi_{\text{lat}}(s, a) = \left\| \sigma(s, a) - D \circ E(\sigma(s, a)) \right\|_2^2.
    \label{eq:phi_lat}
\end{equation}
A high value for $\phi_{\text{lat}}(s, a)$ indicates that the pair $(s, a)$ lies far from the manifold of the training data.

\paragraph{Offline output}

The trained components:
\[
\left\{ E, D, \phi_{\text{lat}}, \hat Q_{\theta}, \sigma(\cdot) \right\}
\]
are exported as a frozen offline artifact, ready to be deployed by the online controller for real-time action selection.

\begin{figure}[h]
    \centering
    \includegraphics[width=0.9\textwidth]{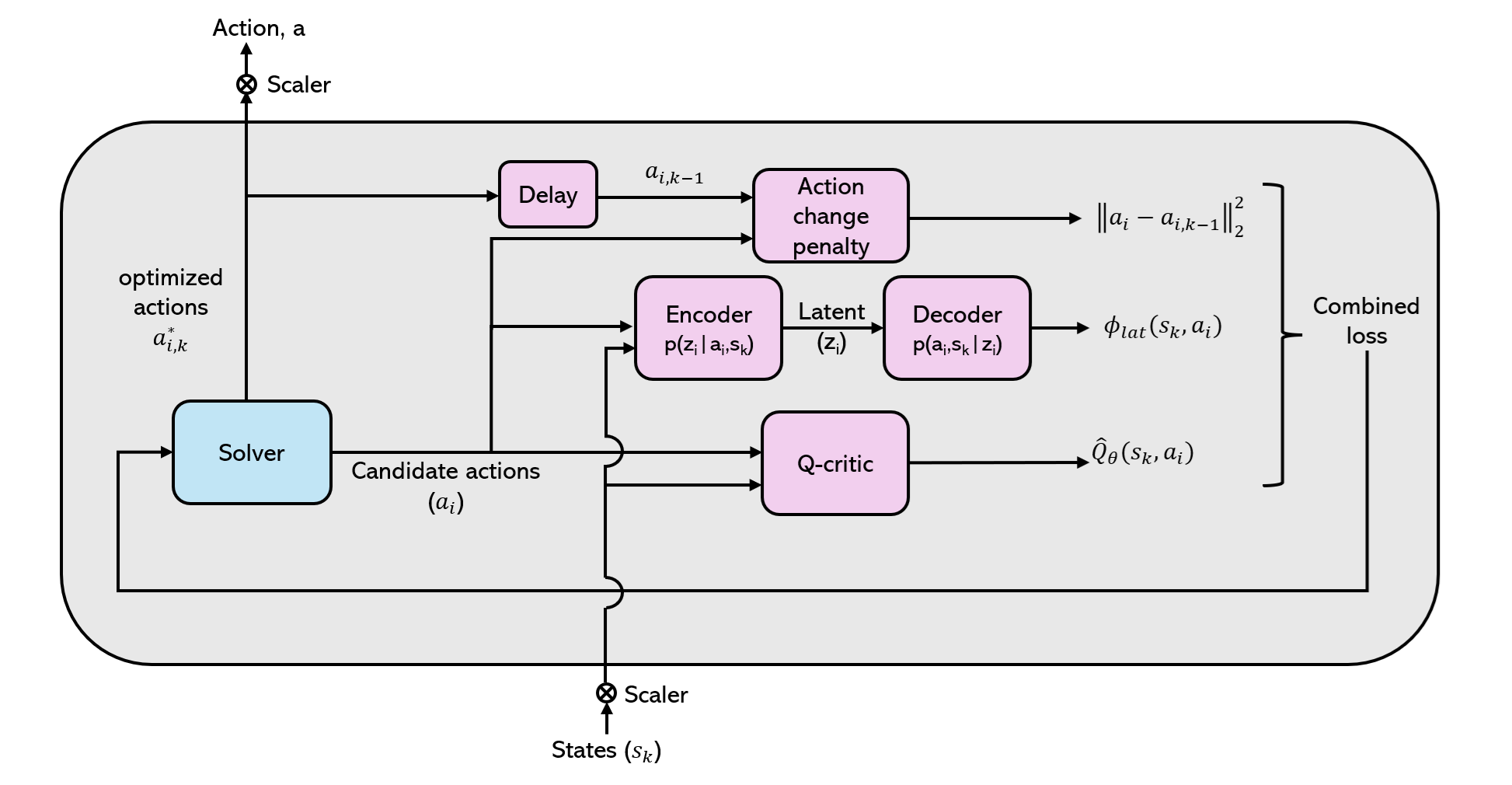}
    \caption{Block diagram representation of the online elements in the HOFLON-RL algorithm.}
    \label{fig:online_block}
\end{figure}

\subsubsection{Online optimization phase}
\label{sec:online}

During online deployment, the trained and frozen models $\{\hat{Q}_{\theta}, E, D\}$ are used by a real-time optimizer to select the next action at each control step $k$. The controller receives the current state $s_k$, computes the error terms $e_k$ and $I_k$, and then solves the following constrained optimization problem to find the optimal action $a_k^{\star}$:
\begin{equation}
    a_k^{\star} = \arg\max_{a \in [0,1]^m} \left[ \hat{Q}_{\theta}(s_k, e_k, I_k, a) - \lambda_1 \phi_{\text{lat}}(s_k, a) - \lambda_2 \|a - a_{k-1}\|_2^2 \right].
    \label{eq:online_opt}
\end{equation}
The objective function balances three critical goals:
\begin{itemize}
    \item \textbf{Value maximization:} The term $\hat{Q}_{\theta}(\cdot)$ drives the policy to select actions that lead to high long-term returns.
    \item \textbf{Safety/distributional constraint:} The penalty $\phi_{\text{lat}}(\cdot)$ ensures the selected action is "in-distribution," preventing the optimizer from exploiting erroneous regions of the Q-function.
    \item \textbf{Control smoothness:} The term $\|a - a_{k-1}\|_2^2$ penalizes large changes in consecutive actions, ensuring smooth actuator movements.
\end{itemize}
The weights $\lambda_1$ and $\lambda_2$ are hyperparameters that tune the trade-off between performance, safety, and smoothness. This optimization problem is solved at each time step using a numerical solver (e.g., Powell's method), as outlined in Algorithm 1 and illustrated in Fig. \ref{fig:online_block}. This formulation translates the rich information learned from offline data into a high-performance, constraint-aware, and stable online control policy.

\begin{algorithm}[h]
\caption{HOFLON-RL}
\KwData{initial observation $x_0$; previous action $a_{-1}$;
        integrator vector $I_0=\mathbf 0$}
\KwIn{weights $(f_1,f_2,f_3)$; sampling period $\Delta t$;
      tolerance $\varepsilon$; corridor length $n_c$}

\For{$k = 0,1,2,\dots$}{
  $x_k \leftarrow$ \textbf{observe} plant\;
  
  $e_k \leftarrow g\!\bigl(x_k,x_{\mathrm{sp}}\bigr)$\;  

  \textit{/* derivative-free optimization */}\;
   $a_k^{\star}\leftarrow$ Powell $J(a),[0,1]^m$\;
 
  \textbf{apply} $a_k^{\star}$; wait $\Delta t$ and receive $x_{k+1}$\;

  \If{$a_k^{\star}$ not clipped}{
      $I_{k+1} \leftarrow I_k + e_k\,\Delta t$\;
  }\Else{
      $I_{k+1} \leftarrow I_k$\;
  }

  \If{$|e_k^{(i)}| < \varepsilon\;\forall i$ for $n_c$ steps}{
      $(f_2,f_3) \leftarrow \texttt{tighten\_weights}()$\;
  }
}
\end{algorithm}

\section{Case studies}\label{sec:dev_cstr}
We assess the performance of the proposed HOFLON-RL approach on two challenging, industrially-motivated simulations that encapsulate common operational hurdles in the process industries. Both scenarios are formulated as Multiple-Input, Multiple-Output (MIMO) control problems characterized by significant cross-couplings between variables, demanding a coordinated control strategy to manage these interactions effectively.

\begin{itemize}
    \item \textbf{Polymerization reactor start-up.} The first case study addresses the critical start-up phase of an exothermic polymerization reactor. Using a Continuous Stirred-Tank Reactor (CSTR) model, the objective is to guide the system from an inert condition to its target steady state. This must be achieved without violating tight operational constraints on reactor temperature and monomer conversion, which is an archetypal safety–critical control problem.

    \item \textbf{Paper-machine grade change.} The second benchmark focuses on economic performance by simulating a product grade change using the multivariable paper-machine model from \citet{kao2025control}. Here, the controller’s task is to efficiently transition multiple quality variables—the final product's basis weight, ash content, and moisture—to new specifications, with the goal of minimizing costly off-spec production.
\end{itemize}

Both case studies are made available at \url{https://github.com/AISL-at-Imperial-College-London} released as Python code for the simulators accompanied by curated offline datasets, the offline training, and online control scripts. The shared materials should enable reproducible evaluation of policy-learning methods on problems that demand precise control across fundamentally different operating regimes.

\subsection{Polymerization reactor start‑up}

This case study uses the suspended, exothermic CSTR model introduced by \citet{maner1997polymerization} to benchmark the HOFLON–RL algorithm on a realistic \emph{start‑up} operation.  The objective is to drive the polymer concentration $C_P$ from~0 to \SI{100}{kg.m^{-3}} while keeping the reactor temperature near \SI{350}{K} in the presence of strong heat release and risk of thermal runaway.

\subsubsection{Polymerization reactor model}
\begin{figure}[htb]
    \centering
    \includegraphics[width=0.6\textwidth]{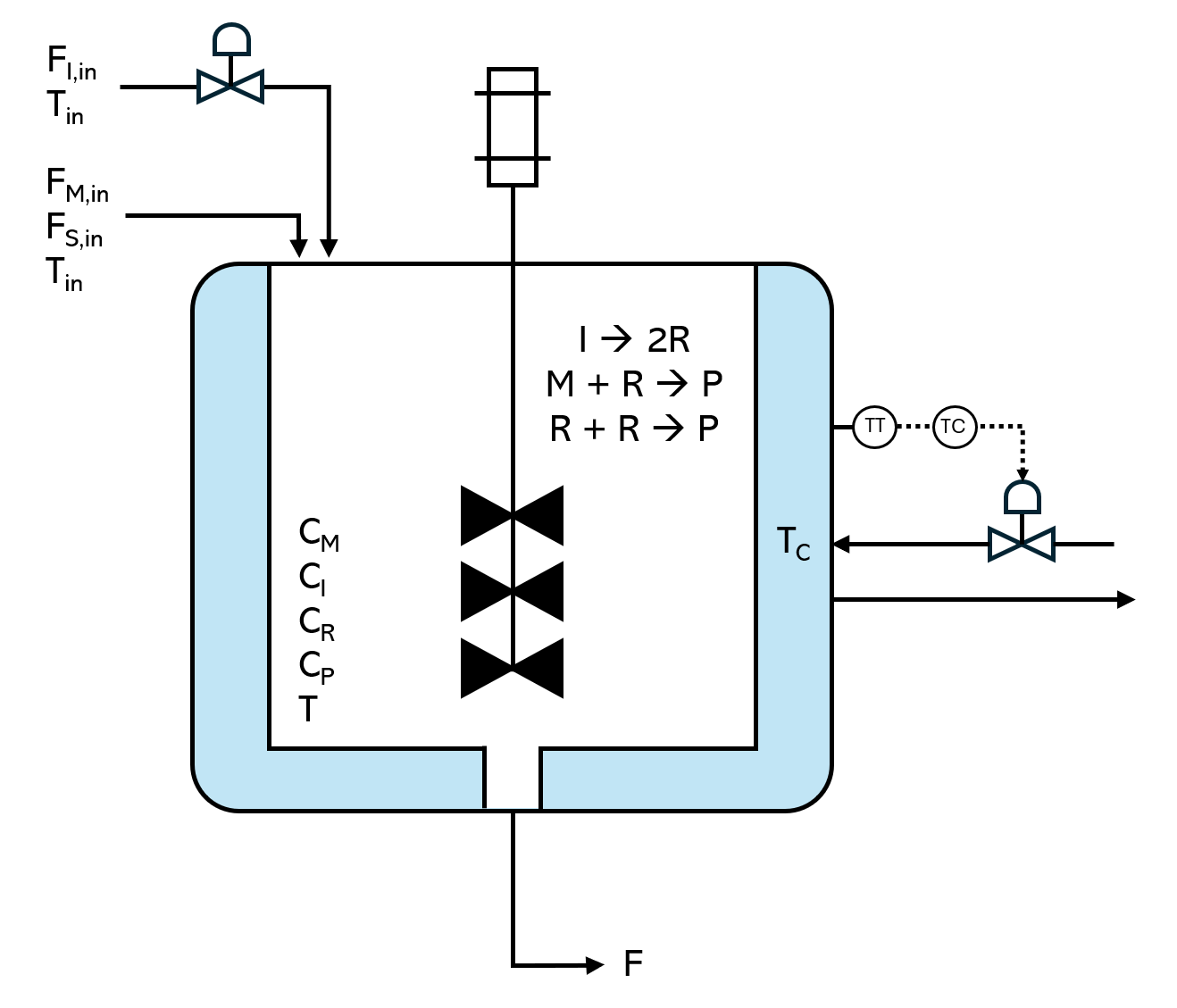}
    \caption{Schematic of the polymerization CSTR environment. Control actions include the initiator feed rate ($F_{\text{M,in}}$) and cooling jacket temperature $T_C$. The reactor states include monomer concentration ($C_M$), initiator concentration ($C_I$), radical concentration ($C_R$), polymer concentration ($C_P$), and reactor temperature ($T$), where the latter two are also the controlled variables.}
    \label{fig:poly_cstr_schematic}
\end{figure}

A schematic of the CSTR is shown in Fig.\ref{fig:poly_cstr_schematic}.
The CSTR is fed with a solvent (S) containing monomer species (M) and an initiator species (I) at temperature $T_\text{in}$. In the CSTR, the monomer and initiator react to form a polymer product (P) via a suspended lumped radical (R) mechanism. The CSTR outlet flowrate is assumed to be equal to the total inlet flowrate, maintaining a constant liquid filled reactor volume (1 m$^3$). Perfect mixing is assumed along with constant fluid density (1000 kg/m$^3$). Any gas phase effects and changes in viscosity due to polymer build-up are neglected.
 
The reaction pathway includes thermal auto-decomposition of the initiator to generate radicals (\ref{reaction:init}), propagation of monomer with radicals to form polymer chains (\ref{reaction:prop}), and bimolecular termination of radicals (\ref{reaction:term}). In this mechanism, all radical species (including free radical and propagating chain radicals) are represented with a single concentration variable for the lumped radicals, $C_R$. This approximation simplifies the kinetic model by avoiding the need to track individual chain lengths. The reaction set (\ref{reaction:init}--\ref{reaction:term}) describes the chemical reactions.
 
\begin{align}
    \text{Initiation:} \quad & \ce{I ->[k_i] 2R{\cdot}} \label{reaction:init}\\
    \text{Propagation:} \quad & \ce{M + R{\cdot} ->[k_p] P{\cdot}} \label{reaction:prop}\\
    \text{Termination:} \quad & \ce{R{\cdot} + R{\cdot} ->[k_t] P} \label{reaction:term}
\end{align}
 
Reaction kinetics are modeled using temperature-dependent Arrhenius expressions with constant pre-exponential factors and activation energies given in Table \ref{tab:poly_cstr_params}. All reactions are assumed to be kinetically controlled with no mass transfer limitations.
 
\begin{equation}
    k = A \exp\left(-\frac{E}{RT}\right)
\end{equation}
 
The reaction rates are defined as follows:
\begin{align}
    \text{Initiation:} \quad & r_i = k_i C_I \\
    \text{Propagation:} \quad & r_p = k_p C_M C_R \\
    \text{Termination:} \quad & r_t = k_t C_R^2
\end{align}
 
\subsubsection{Dynamics and numerical integration}
The reactor dynamics are governed by a system of coupled ordinary differential equations (\ref{ode1}--\ref{ode5}) using the parameters in Table \ref{tab:poly_cstr_params}. Heat effects are explicitly modeled, with an exothermic heat of reaction and thermal exchange with a cooling jacket. These are solved at each simulation step using SciPy's solveivp function with a stiff solver (BDF). The integration time step is set to 30 minutes to reflect typical control intervals for large-scale batch or semi-batch reactors. The simulation is terminated after 100 hours of operation unless otherwise specified.
 
\begin{align}
    \frac{dC_M}{dt} &= \frac{F}{V}\left(C^\text{in}_M - C_M\right) - r_p \label{ode1}\\
    \frac{dC_I}{dt} &= \frac{F}{V}\left(C^\text{in}_I - C_I\right) - r_i \\
    \frac{dC_R}{dt} &= \frac{F}{V}\left(C^\text{in}_R - C_R\right) + 2r_i - 2r_t \\
    \frac{dC_P}{dt} &= \frac{F}{V}\left(C^\text{in}_P - C_P\right) + r_p \\
    \frac{dT}{dt} &= \frac{F}{V}(T_f - T) - \frac{U A (T - T_c)}{\rho C_p V} - \frac{r_p \Delta H_{\text{rxn}}}{\rho C_p} \label{ode5}
\end{align}
 
Table \ref{tab:poly_cstr_params} gives the parameters used to construct the polymerization reactor environment. The reaction is inspired by the polymerization of vinyl acetate in benzene as presented in \citep{maner1997polymerization}, with some simplifications and rounding of parameters.
 
\begin{table}[htb]
    \centering
    \caption{Summary of reaction / reactor parameters used in the PolyCSTR environment}
    \label{tab:poly_cstr_params}
    \begin{tabular}{llc}
        \hline
        \textbf{Category} & \textbf{Parameter} & \textbf{Value} \\
        \hline
        \multirow{5}{*}{Physical Constants}
        & Gas constant $R$ & 8.314 J/mol/K \\
        & Density $\rho$ & 1000 kg/m$^3$ \\
        & Heat capacity $C_p$ & 2000 J/kg/K \\
        & Heat transfer coefficient $U$ & 500 J/s/K \\
        & Heat of reaction $\Delta H_{\text{rxn}}$ & -100,000 J/mol \\
        \hline
        \multirow{5}{*}{Reactor Configuration}
        & Volume $V$ & 1.0 m$^3$ \\
        & Feed solvent rate $F_S$ & 80 kg/h \\
        & Feed monomer rate $F_M$ & 100 kg/h \\
        & Feed temperature $T_f$ & 350 K \\
        \hline
        \multirow{6}{*}{Kinetics}
        & $A_{\text{init}}$ & $1 \times 10^9$ 1/s \\
        & $E_{\text{init}}$ & 125,000 J/mol \\
        & $A_{\text{prop}}$ & $4 \times 10^4$ m$^3$/mol/s \\
        & $E_{\text{prop}}$ & 25,000 J/mol \\
        & $A_{\text{term}}$ & $1 \times 10^6$ m$^3$/mol/s \\
        & $E_{\text{term}}$ & 15,000 J/mol \\
        \hline
        \multirow{2}{*}{Control Limits}
        & Initiator feed rate & [0.0, 2.5] kg/h \\
        & Coolant temperature & [300, 355] K \\
        \hline
        \multirow{2}{*}{Simulation Settings}
        & Time step & 1800 s \\
        & Max simulation time & 100 hours \\
        \hline
    \end{tabular}
\end{table}
\subsubsection{RL observation and action spaces}
\label{sec:poly_cstr_spaces}
\paragraph{State scaling.}  Concentrations are divided by \num{100}~(\si{kg.m^{-3}}) and temperature by \SI{350}{K}, yielding dimensionless variables in roughly the range $[0,2]$: $P_s = C_P/100$ and $T_s = T/350$.

\paragraph{Actions.}  The agent selects two dimensionless inputs in $[0,1]$:
\begin{align*}
    u_I   &= \frac{f_{I} - f_{I,\text{min}}}{f_{I,\text{max}}-f_{I,\text{min}}}, &
    u_{T_c} &= \frac{T_c - T_{c,\text{min}}}{T_{c,\text{max}}-T_{c,\text{min}}},
\end{align*}
with $f_I\in[0,2.5]$~kg\,h$^{-1}$ and $T_c\in[280,400]$~K.

\paragraph{Observations.}  The agent receives the five scaled states with Gaussian measurement noise ($1\,\sigma$ standard deviations $\sigma=[0.02,0,0,0.005,0.00057]$).

\subsubsection{Data‑generation methodology}
\label{sec:poly_cstr_data}
Offline datasets are produced by a two‑loop PI controller operating in the \emph{scaled} domain.

\begin{itemize}
    \item \textbf{Controller gains.}  Base gains $\bigl(K_{p,P}^{\*},K_{i,P}^{\*},K_{p,T}^{\*},K_{i,T}^{\*}\bigr)=\bigl(1.04,\,8\times10^{-5},\,0.012,\,3.8\times10^{-5}\bigr)$ are multiplied by independent random factors sampled uniformly from $[0.3,1.0]$ at the start of each run.
    \item \textbf{Anti‑wind‑up.}  Integral terms are frozen whenever a manipulated variable saturates.
    \item \textbf{Scheduling.}  Each run simulates \SI{100}{h} with a \SI{30}{min} control interval, yielding 200 logged points.  We generate 100 statistically independent runs, giving \num{20\,000} transitions in the final dataset.
    \item \textbf{Measurement noise.}  Additive Gaussian noise with the standard deviations given in Section~\ref{sec:poly_cstr_spaces} is added to the state vector before it is fed to the controller and recorded in the dataset.
\end{itemize}

\paragraph{Pairing.}
A diagonal two-loop PI structure is used: the polymer-concentration error $e_P$ actuates the initiator feed $u_I$, and the temperature error $e_T$ actuates the coolant set-point $u_{T_c}$; no cross terms are applied.

\subsubsection{Reward function}
\label{sec:poly_cstr_reward}
Let $P_s$ and $T_s$ denote the \emph{scaled} polymer concentration and temperature, with set‑points $P^{\text{sp}}=1$ and $T_s^{\text{sp}}=1$.  Define the instantaneous squared error
\[
    e(k)=\bigl(P_s(k)-P^{\text{sp}}\bigr)^2 + \bigl(T_s(k)-T_s^{\text{sp}}\bigr)^2
\]
and the improvement $\Delta e(k)=e(k-1)-e(k)$.  For actions $u(k)=[u_I(k),u_{T_c}(k)]$ and step changes $\Delta u = u(k)-u(k-1)$, the reward has the form:
\begin{equation}\label{eq:poly_reward}
    r(k)=\alpha\Bigl[-e(k)+\exp\!\Bigl(-\tfrac{e(k)}{\sigma^2}\Bigr)\Bigr] + \Delta e(k) - \lambda_1\,\Delta u_I^2 - \lambda_2\,\Delta u_{T_c}^2,
\end{equation}
with $\sigma=0.05$.  The first bracketed term rewards proximity to the set‑point, the second rewards progress, and the last term discourages aggressive control moves.

\subsubsection{Logged dataset}
\label{sec:poly_cstr_logged}

At each 30-minute interval, the driver appends one row containing:
identifiers (\texttt{run\_id}, time (min));
scaled states $M,I,R,P,T_s$;
scaled errors \texttt{err\_P}, \texttt{err\_T}, \texttt{int\_err\_P}, \texttt{int\_err\_T};
scaled MVs \texttt{u\_I}, \texttt{u\_Tc};
and the reward.

\subsubsection{Control scenario}
\label{sec:poly_cstr_scenarios}
Only the \textbf{start‑up} scenario is considered:
\begin{itemize}
    \item \emph{Warm start.}  The reactor initially contains solvent and monomer at $T=\SI{350}{K}$ with \emph{no} initiator ($u_I=0$), hence $P_s(0)=0$.  The task is to reach $P_s=1$ and maintain $T_s=1$ without triggering thermal runaway.
\end{itemize}

The combination of strong exothermic behavior, nonlinear kinetics, measurement noise and randomly perturbed PID loops results in a dataset that contains both cautious and aggressive trajectories—making it an informative benchmark for offline policy learning.

\subsection{Paper machine grade change}

This case study examines the dynamic modeling of a grade change operation in a paper machine, based on the work by \citet{ko2003modeling}. A grade change involves transitioning the production process from one paper specification to another, a complex procedure that can lead to significant amounts of off‑specification product if not carried out precisely. The objective is to minimize this transition time and maintain product quality.

\begin{figure}[htb]
    \centering
    \includegraphics[width=0.95\textwidth]{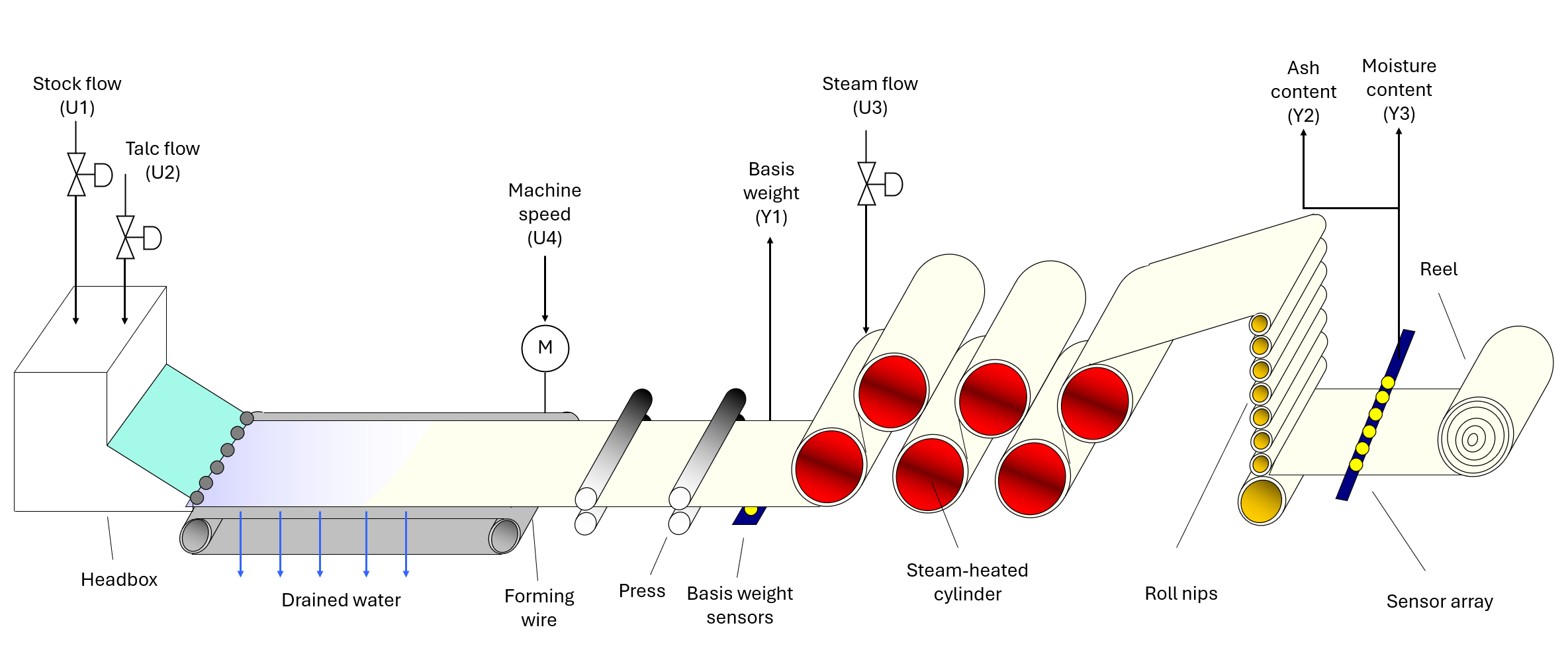}
    \caption{Schematic of the paper machine indicating the manipulated variables (or actions) and the observed process outputs.}
    \label{fig:paper_machine_schematic}
\end{figure}

The physical system encompasses the process from the headbox through to the main dryer section and is illustrated in Fig.\ref{fig:paper_machine_schematic}. The key process variables are:
\begin{itemize}
    \item \textbf{Manipulated Variables (Inputs or actions):} Stock flow, filler (clay) flow, steam pressure, and machine speed.
    \item \textbf{Controlled Variables (Outputs or observations):} Basis weight, ash content, and moisture content.
\end{itemize}

In \citet{ko2003modeling}, closed-loop system identification is used to obtain a discrete‑time, linear time‑invariant (LTI) state‑space model. The model is defined by the state‑space matrices $A$, $B$, $C$, and the direct feedthrough matrix $D$, which is a zero matrix as is common for processes with inherent delays between inputs and outputs.

The state‑space model is given by
\begin{align}
    x(k+1) &= A\,x(k) + B\,u(k), \label{eq:state} \\
    y(k)   &= C\,x(k) + D\,u(k). \label{eq:output}
\end{align}
Here, $x(k)$ is the state vector, $u(k)$ is the vector of manipulated‑variable deviations, and $y(k)$ is the vector of controlled‑variable deviations.

The identified matrices are:

\paragraph{State matrix $A$}
\[
A = \begin{bmatrix}
 0.676 &  0.349 & -0.153 & -2.462 & -2.182 &  0.735 & -0.477 & -1.326 &  0.201 & -0.199 \\
-1.309 &  2.494 & -0.601 & -10.70 & -3.071 & -0.850 & -0.803 & -3.458 &  0.641 & -0.372 \\
 0.941 & -1.215 &  1.472 &  9.160 &  4.575 & -3.024 &  1.526 &  4.882 & -1.176 &  0.902 \\
-0.305 &  0.362 & -0.132 & -1.566 & -0.493 & -0.145 & -0.079 & -0.553 &  0.151 & -0.025 \\
-0.043 &  0.077 & -0.029 & -0.612 &  0.619 &  0.449 &  0.164 & -0.017 & -0.098 &  0.035 \\
-0.318 &  0.358 & -0.108 & -2.318 &  0.066 &  0.193 & -0.523 & -0.131 &  0.146 &  0.064 \\
-0.027 &  0.071 & -0.021 & -0.491 &  0.052 &  0.398 &  0.542 & -0.111 &  0.434 & -0.060 \\
 0.179 & -0.203 &  0.024 &  1.140 & -0.029 & -0.006 & -0.046 & -0.143 &  0.162 &  0.023 \\
 0.437 & -0.511 &  0.161 &  3.481 &  0.593 &  0.311 & -0.526 & -0.215 &  0.584 &  0.220 \\
 0.420 & -0.519 &  0.174 &  3.604 &  0.577 &  0.241 &  0.102 &  0.272 & -0.462 &  0.354
\end{bmatrix}
\]

\paragraph{Input matrix $B$}
\[
B = \begin{bmatrix}
-5.15\times10^{-3} & -1.977\times10^{-2} &  7.755\times10^{-1} & -1.381\times10^{-2} \\
-1.938\times10^{-2} & -5.644\times10^{-2} &  3.630 & -5.918\times10^{-2} \\
-1.642\times10^{-2} &  3.865\times10^{-2} & -5.436\times10^{-1} &  5.292\times10^{-2} \\
-2.95\times10^{-3} & -7.79\times10^{-3} &  4.429\times10^{-1} & -1.406\times10^{-2} \\
 7.13\times10^{-3} & -8.50\times10^{-4} & -6.990\times10^{-1} & -4.00\times10^{-3} \\
-1.25\times10^{-2} &  3.40\times10^{-4} &  8.226\times10^{-1} & -1.156\times10^{-2} \\
 5.65\times10^{-3} &  1.84\times10^{-3} & -2.227\times10^{-1} & -3.76\times10^{-3} \\
 7.23\times10^{-3} & -1.536\times10^{-2} &  9.547\times10^{-1} &  4.46\times10^{-3} \\
 6.13\times10^{-3} &  8.90\times10^{-4} &  2.952\times10^{-1} &  1.798\times10^{-2} \\
 1.09\times10^{-3} &  4.58\times10^{-3} &  4.471\times10^{-2} &  1.986\times10^{-2}
\end{bmatrix}
\]

\paragraph{Output matrix $C$}
\[
C = \begin{bmatrix}
-1.742 &  0.390 &  0.118 &  2.347 & -2.043 &  1.614 & -0.316 &  0.125 &  0.037 & -0.141 \\
-0.304 & -0.088 & -0.159 & -0.141 & -0.419 & -0.070 &  0.064 & -0.355 & -0.106 &  0.132 \\
-0.207 &  0.189 &  0.002 & -2.854 & -0.934 &  0.275 & -0.256 & -0.822 &  0.159 & -0.186
\end{bmatrix}
\]

\paragraph{Feedthrough matrix $D$}
\[
D = \begin{bmatrix}
0 & 0 & 0 & 0 \\
0 & 0 & 0 & 0 \\
0 & 0 & 0 & 0
\end{bmatrix}
\]

Even though \citet{ko2003modeling} does not state a sampling rate explicitly, considering other works from the same group such as \citet{yeo2005model}, we can estimate the sampling rate to be $30\,\si{\second}$

\subsubsection{Data generation methodology}

To simulate the system for controller development and testing, a data generation script was created. This script simulates the closed‑loop response of the paper machine model for the grade change under PI control, introducing variability to create a rich dataset. The entire simulation, including the control law, operates on scaled data to normalize the variable ranges.

\paragraph{Scaling}

All variables are scaled to facilitate control and learning.
\\
\newline
\textbf{Controlled variables (CVs)} – basis weight ($Y_1$), ash content ($Y_2$) and
moisture content ($Y_3$) – are expressed in
\si{g.m^{-2}}, \si{\percent}, and \si{\percent}, respectively.
They are first converted to deviations from nominal values,
$Y_{\text{nom}}=[64.0\,\si{g.m^{-2}},\,7.0\,\si{\percent},\,9.0\,\si{\percent}]^{\mathsf T}$,
and then normalised:
\begin{align}
    y_{\text{scaled}} &= \frac{y_{\text{phys}} - Y_{\text{nom}}}{Y_{\text{nom}}}, \\
    y_{\text{phys}}   &= Y_{\text{nom}}\bigl(1 + y_{\text{scaled}}\bigr).
\end{align}
\\
\newline
\textbf{Manipulated variables (MVs)} are scaled to the range $[-1,1]$ using
the physical limits in Table~\ref{tab:mv_limits}.

\begin{table}[htb]
    \centering
    \caption{Physical limits used for MV scaling}
    \label{tab:mv_limits}
    \sisetup{table-number-alignment=center}
    \begin{tabular}{lcc}
        \hline
        \textbf{MV} & {$u_{\min}$} & {$u_{\max}$} \\
        \hline
        Stock flow (\si{L\,min^{-1}})            & 300  & 600  \\
        Filler flow (\si{L\,min^{-1}})           &   0  & 200  \\
        Steam pressure (\si{bar})                &   1  &   3  \\
        Machine speed (\si{m\,min^{-1}})         & 800  & 1000 \\
        \hline
    \end{tabular}
\end{table}

The forward and inverse maps are
\begin{align}
    u_{\text{scaled}} &= 2\,\frac{u_{\text{phys}} - u_{\min}}{u_{\max} - u_{\min}} - 1, \\
    u_{\text{phys}}   &= u_{\min} + 0.5\,(u_{\max} - u_{\min})\bigl(u_{\text{scaled}} + 1\bigr).
\end{align}

\paragraph{Simulation and control loop:}
The script executes 100 simulation runs, each for 250 time steps.
\\
\newline
\textbf{Initialization:} Each run begins with a random initial state $x(0)$ drawn from a uniform distribution between $-0.1$ and $0.1$.
\\
\newline
\textbf{PI controller with gain uncertainty:} A PI controller is used to drive the system to the desired set‑point. To ensure the data reflects realistic process variations, the controller gains ($K_p$, $K_i$) are randomized in each run. Base gains are multiplied by random factors drawn from specified uncertainty bands.

The control action $u_{\text{scaled}}$ is calculated based on the scaled error $e_{\text{scaled}} = Y_{\text{sp,scaled}} - y_{\text{scaled}}$:
\[
    u_{\text{scaled}}(k) = u_{\text{scaled,nom}} + K_p \cdot e_{\text{loop}}(k) + \text{int\_term}(k).
\]
Here, $e_{\text{loop}}$ is a four‑element error vector (errors for $Y_1$, $Y_2$, $Y_3$, and $Y_1$ again) used to compute the four MV actions.
\\
\newline
\textbf{Anti‑windup:} An anti‑windup mechanism is included. The integral term is only updated if the controller output is not saturated and the error is driving it further into saturation.
\\
\newline
\textbf{State update:} The plant state is updated at each step. The control law computes the scaled MV $u_{\text{scaled}}$, which is then descaled to its physical value $u_{\text{phys}}$. The deviation from the nominal operating point, $\Delta u_{\text{phys}} = u_{\text{phys}} - U_{\text{nom}}$, is used in the state‑space equation
\[
    x(k+1) = A\,x(k) + B\,\Delta u_{\text{phys}}(k).
\]
The resulting state deviation is used to calculate the physical CV output, to which multiplicative noise is added before it is re‑scaled for the controller.
\paragraph{Pairing.}
A diagonal four-loop PI structure is used with second basis-weight loop in a SIMO control arrangement, where stock flow $U_1$ and machine speed $U_4$ are driven by the basis-weight error $e_{Y_1}$; filler (clay) flow $U_2$ is driven by the ash-content error $e_{Y_2}$; and steam-header pressure $U_3$ is driven by the moisture error $e_{Y_3}$.

\subsubsection{Reward function}
\label{sec:pm_reward}

Let $e_i(k)$ be the \emph{scaled} tracking error of the $i$-th controlled
variable considering the setpoints corresponding to the ongoing grade change scenario at step $k$, and define the instantaneous aggregate error

\[
    e(k)=\sum_{i=1}^{3} e_i^{\,2}(k).
\]

Its one-step improvement is
$\Delta e(k)=e(k-1)-e(k)$. For the action vector
$u(k)=[u_1(k),u_2(k),u_3(k),u_4(k)]$ (scaled stock flow, filler flow, steam
pressure and machine speed) and the step change
$\Delta u(k)=u(k)-u(k-1)$, the reward is

\begin{equation}\label{eq:pm_reward}
    r(k)=\alpha\Bigl[-e(k)+\exp\!\Bigl(-\tfrac{e(k)}{\sigma^2}\Bigr)\Bigr]
          +\Delta e(k)
          -\lambda\,\lVert\Delta u(k)\rVert_2^{2},
\end{equation}

with $\sigma=0.1$. As in the previous case study, the first
bracketed term rewards proximity to the set-point, the middle term rewards
progress, and the last term discourages aggressive control moves.

\subsubsection{Logged dataset}
\label{sec:pm_logged}

At each control interval, the simulator appends one row containing:
identifiers (run, step); scaled CVs $\mathrm{Y1},\mathrm{Y2},\mathrm{Y3}$ (basis weight, ash, moisture);
scaled MVs $\mathrm{U1}\!-\!\mathrm{U4}$ (stock flow, filler flow, steam-header pressure, machine speed);
error channels $e_1\!-\!e_3$ and \texttt{err\_metric} ($\sum_i e_i^2$);
integrator states \texttt{int1}\dots\texttt{int4}; and the reward.
All values are normalized (CVs, MVs) or dimensionless, so the CSV is ready for direct use by RL algorithms without further scaling.

\subsubsection{Control scenario}

The original study by Ko et al. (2003) identified seven distinct paper grades, labeled A through G, each defined by specific setpoints for the key quality variables: basis weight, ash content, and moisture content, as detailed in Table~\ref{tab:grade_changes}. For the control problem in this work, we focus on a single, challenging transition: a grade change from Grade A to Grade D. This scenario requires the controller to steer the process from a nominal state with a basis weight of 64 g/m$^2$ and an ash content of 7\% to a new target state with a basis weight of 102 g/m$^2$ and an ash content of 17\%. The moisture content setpoint remains constant at 9\% for both grades. This represents a significant change in operating conditions, demanding a coordinated response from all four manipulated variables to manage the transition efficiently.

\begin{table}[htb]
    \centering
    \caption{Set point changes according to paper grades \citep{kao2025control}}
    \label{tab:grade_changes}
    \begin{tabular}{lccccccc}
        \hline
        \textbf{Paper grade} & \textbf{A} & \textbf{B} & \textbf{C} & \textbf{D} & \textbf{E} & \textbf{F} & \textbf{G} \\
        \hline
        Basis weight (g/m$^2$) & 64 & 49 & 78 & 102 & 63 & 100 & 78.5 \\
        Ash content (\%)       & 7  & 7  & 17 & 17  & 9  & 17 &  14    \\
        Moisture content (\%)  & 9  & 8  & 9  & 9   & 9  & 9    & 9    \\
        \hline
    \end{tabular}
\end{table}
 
\section{Results} \label{sec:results}
 
In this section, we report the empirical performance of HOFLON-RL on the two benchmark problems introduced earlier. We begin with the polymerization-reactor start-up, detailing the specific network architecture, hyper-parameters and training protocol used for each HOFLON-RL formulation, then present the results that highlight the tracking performance. The second part of the section applies the same evaluation to the paper-machine grade-change problem, allowing a comparison of across the two  different MIMO problems. IQL baseline design and the results using the IQL agents are provided for comparison for both case studies.

\subsection{Polymerization reactor start-up} 

Fig. \ref{fig:train_data_poly} depicts 100 simulated episodes of the polymerization-CSTR, each driven by a PI controller whose gains were sampled at random. The trajectories span the full range of closed-loop behavior observed in the start-up scenario: some runs settle the polymer concentration and temperature rapidly at their set-points, whereas others display oscillations or sluggish convergence caused by less favorable tuning. Such variability in the data is important for offline RL because it supplies a rich collection of state–action pairs from both well-regulated and poorly regulated regimes. Exposing the learning agent to this broad operating envelope—even those episodes that verge on unsafe—helps foster policies that are robust and, importantly, safe across diverse process conditions.
 
\begin{figure}[t]
    \centering
    \includegraphics[width=\textwidth]{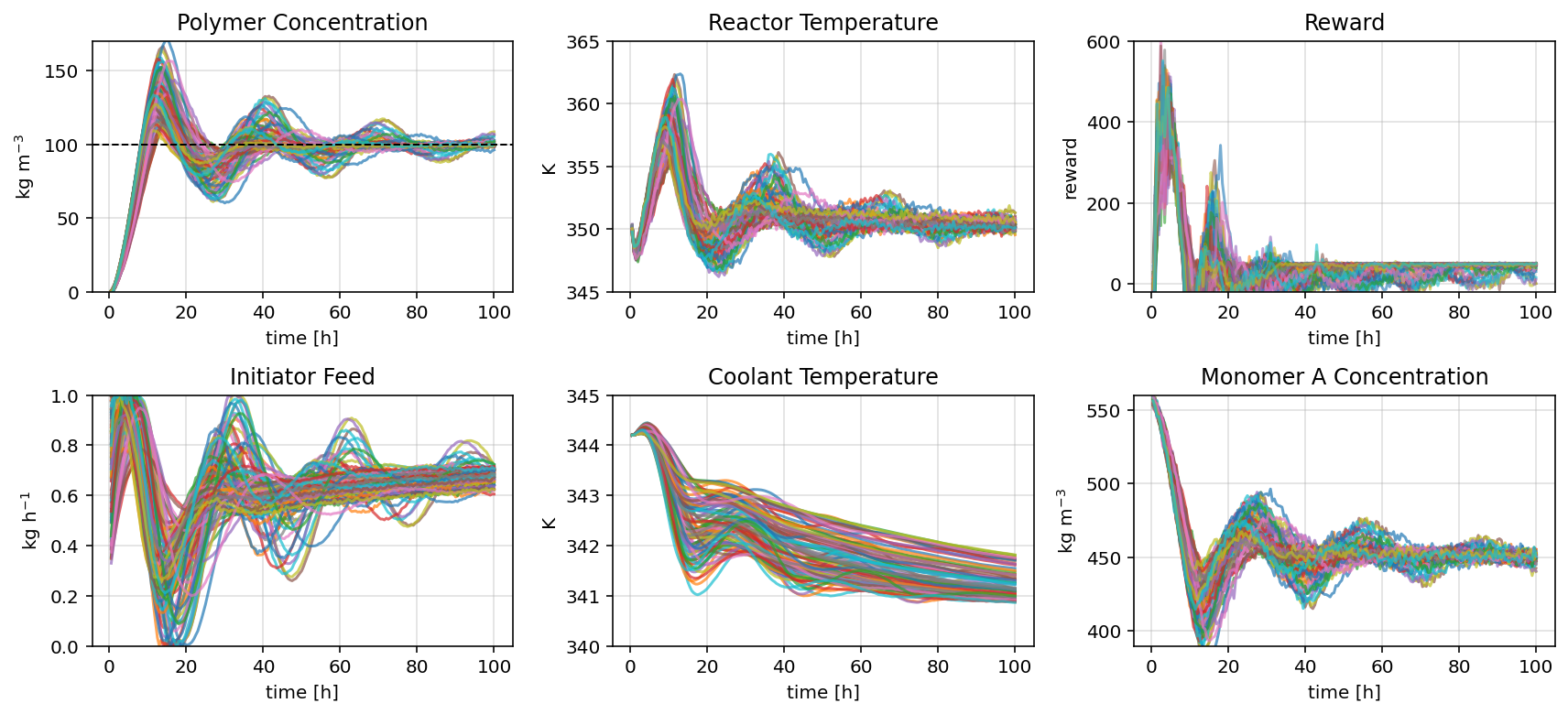}
    \caption{100 PI controlled episodes from the polymerization CSTR start-up scenario.}
    \label{fig:train_data_poly}
\end{figure}
 
\subsubsection{HOFLON: offline learning (polymerization)}
\label{sec:offline_stage}

Offline training consists of two parts: (i) fitting a value function
\(Q_\theta(s,a)\) on the historical returns and (ii) learning a
low-dimensional manifold of feasible state–action pairs with an
adversarial auto-encoder (AAE).  The resulting artifacts are frozen and
used by the online optimizer only for inference.

\begin{table}[htb]
\centering
\caption{Design parameters for the XGBoost Q-critic}
\label{tab:q_params}
\begin{tabular}{lc}
\hline
Feature vector size & 11 (9 state + 2 action) \\
Discount factor & $\gamma = 0.90$ \\
Model type & XGBRegressor \\
Trees / depth & 1600 / 8 \\
Learning rate & 0.03 \\
$\ell_2$ regularisation & $\lambda = 1$ \\
Train/validation split & 80 / 20\,\% \\
Hold-out $R^2$ & 0.96 \\
\hline
\end{tabular}
\end{table}

\paragraph{Q-critic.}
The action-value function, or $Q$-critic, is approximated using an XGBoost regressor. The model's purpose is to estimate the expected discounted future return for any given state-action pair. The input to the regressor is a single feature vector formed by concatenating the nine-element state observation $\bigl[M,I,R,P,T_s,e_P,e_T,i_P,i_T\bigr]$ with the two-element action vector $[u_I,u_{T_c}]$.

\begin{figure}[t]
\centering
\includegraphics[width=0.8\textwidth]{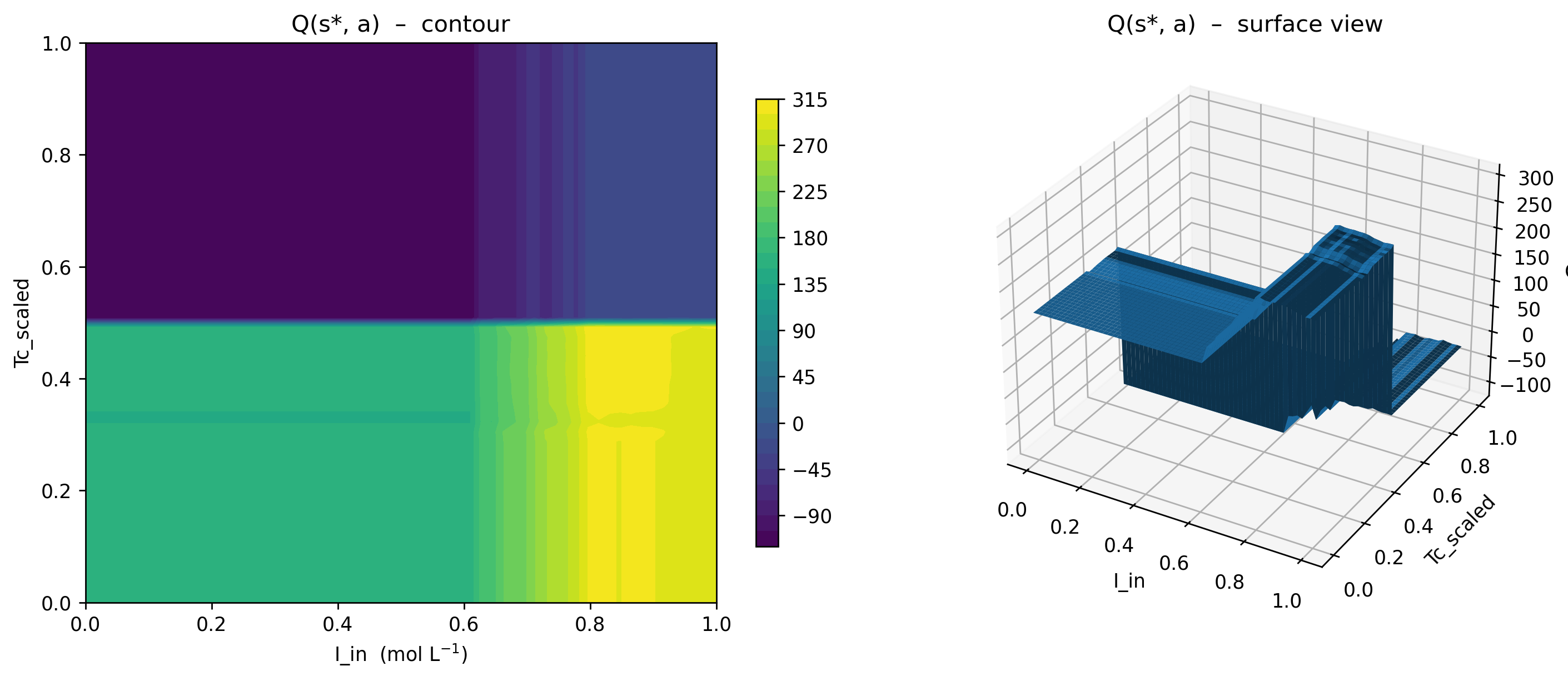}
\caption{The Q-critic surface over the action space evaluated at a state observation from the historical dataset.}
\label{fig:Q-critic}
\end{figure}

The key hyperparameters for the model, such as the number of trees and learning rate, are summarized in Table~\ref{tab:q_params}. Following training, the model showed strong generalization, achieving a coefficient of determination ($R^2$) of 0.96 on a 20\,\% hold-out validation set. A visualization of the learned $Q$-function is presented in Figure~\ref{fig:Q-critic}. The plot shows the critic's output over the entire action space for a fixed, representative state. The resulting surface exhibits piecewise constant and highly nonlinear features expected from an XGBoost model.

\begin{table}[htb]
\centering
\caption{Network architecture and training settings for the AAE}
\label{tab:aae_arch}
\begin{tabular}{@{}llcc@{}} 
\toprule
\textbf{Component} & \textbf{Layer} & \textbf{Output Dim.} & \textbf{Activation} \\
\midrule
\bfseries Encoder & Input & 11 & -- \\
& FC + BN & 64 & LeakyReLU(0.1) \\
& FC & 32 & LeakyReLU(0.1) \\
& FC (latent $z$) & 4 & -- \\
\addlinespace
\bfseries Decoder & FC & 32 & Tanh \\
& FC & 64 & Tanh \\
& FC & 11 & Tanh \\
\addlinespace
\bfseries Discriminator & FC & 32 & LeakyReLU(0.2) \\
& Dropout & -- & $p=0.3$ \\
& FC & 16 & LeakyReLU(0.2) \\
& FC & 1 & Sigmoid \\
\midrule[\heavyrulewidth] 
\multicolumn{2}{@{}l}{\bfseries Hyperparameter} & \multicolumn{2}{l}{\bfseries Value} \\
\midrule
\multicolumn{2}{@{}l}{Latent prior} & \multicolumn{2}{l}{$\mathcal{N}(0, I_4)$} \\
\multicolumn{2}{@{}l}{Optimiser} & \multicolumn{2}{l}{Adam $(\alpha=10^{-5}, \beta_1=0.5, \beta_2=0.999)$} \\
\multicolumn{2}{@{}l}{Batch size} & \multicolumn{2}{l}{256} \\
\multicolumn{2}{@{}l}{Epochs} & \multicolumn{2}{l}{250 (patience 20)} \\
\multicolumn{2}{@{}l}{WGAN Gradient Penalty} & \multicolumn{2}{l}{10} \\
\multicolumn{2}{@{}l}{Test MSE} & \multicolumn{2}{l}{$1.6 \times 10^{-3}$} \\
\bottomrule
\end{tabular}
\end{table}

\paragraph{Adversarial auto-encoder.}
To create a compact and regularized generative manifold of the system data, we employ an Adversarial Autoencoder (AAE). The AAE comprises an encoder, a decoder, and an adversarial discriminator, whose specific network architectures and training settings are detailed in Table~\ref{tab:aae_arch}. All input features are first min-max scaled to the range $[-1, 1]$ to match the decoder's $\tanh$ output activation. The discriminator's role is to force the distribution of the four-dimensional latent code from the encoder to match a standard normal prior, $\mathcal{N}(0, I_4)$. This adversarial training disentangles the latent space, making it smooth and continuous—a vital property for effective sampling during online optimization. The trained AAE using the trajectories from the first 85 runs demonstrates high fidelity in reconstructing the data manifold, achieving a low mean squared error (MSE) of $1.6 \times 10^{-3}$ on the held-out runs. This performance is visually confirmed in Fig.~\ref{fig:aae_recon}, which overlays an original data trajectory with its reconstruction, showing good agreement.

Together, the well-fitted XGBoost critic and the latent AAE manifold model provide the online layer with (i) a dense estimate of long-term value gradients and (ii) a data-supported constraint that penalizes excursions into regions not seen in historical operation.

\begin{figure}[t]
\centering
\includegraphics[width=\textwidth]{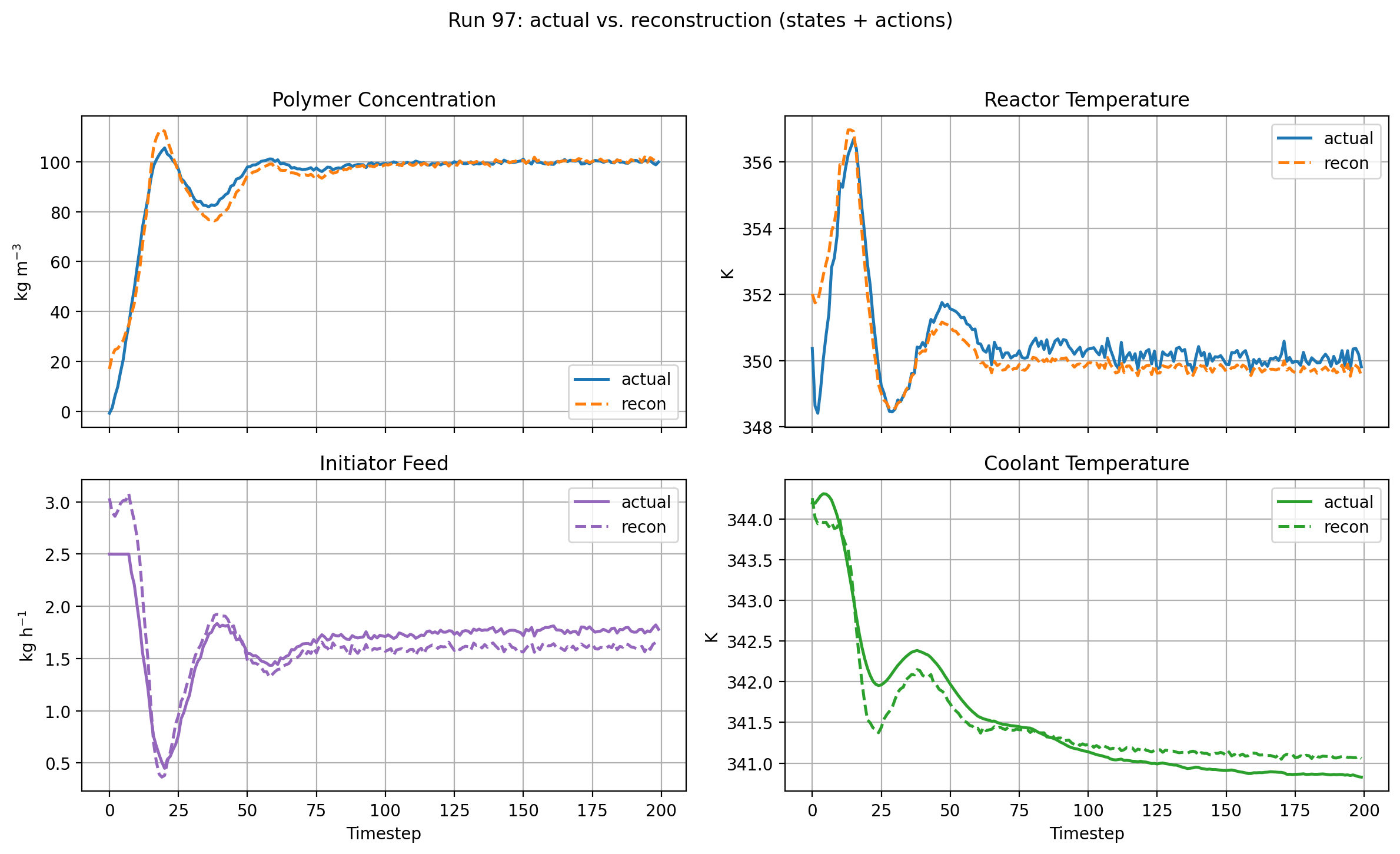}
\caption{Reconstruction of a held-out run using the trained AAE for the controlled variables and the actions.}
\label{fig:aae_recon}
\end{figure}

\subsubsection{HOFLON: online optimization (polymerization)}
\label{sec:online_stage}
\label{sec:online-ho-flon}

\paragraph{Single–step objective.}
At each sampling instant the controller solves the box-constrained optimization given in \ref{eq:online_opt}
The weights $\lambda_1$ and $\lambda_2$ are chosen as 0.08 and 0.04 respectively.

\paragraph{Solution of the optimization problem.}
\label{sec:solver-portfolio}
Because the objective is non-differentiable but cheap to evaluate the following derivative-free solvers are launched sequentially:

\begin{enumerate}
  \item \textbf{Nelder--Mead simplex}.   
        A local simplex that explores a smooth neighborhood extremely
        quickly.  It lacks native bound handling, so its raw optimum is clipped
        back to $[0,1]^2$ before $J$ the objective is re-evaluated.
  \item \textbf{Powell’s conjugate-direction search}.  
        Another local pattern search that \emph{does} enforce bounds and often
        escapes the simplex if Nelder--Mead stalls in a flat valley.
  \item \textbf{DIRECT global pattern search}  
        Provides a lightweight, deterministic global scan of the unit square.
\end{enumerate}

After the three calls the controller re-computes the objective for the \emph{clipped} candidates and applies whichever action set yields the lowest cost.
The simulations were performed on a laptop equipped with an \textbf{Intel\,Core\,Ultra\,9\,185H} processor
(16 physical cores, 22 hardware threads, base-clock 2.5\,GHz). The mean wall-clock per call was 239\,ms (median 233\,ms, maximum 443\,ms), which is comfortably below the sampling period used in the case study. All solvers were warm-started from the previous control moves.

\subsubsection{IQL agent training (polymerization)}

\paragraph{Value function (expectiles).}
We first fit an explicit value function \(V_\tau(s)\) by expectile regression on Monte-Carlo returns \(G_t=\sum_{k\ge 0}\gamma^{k} r_{t+k}\). Expectiles are obtained via iterative reweighted least squares with asymmetric squared-loss weights \(\rho_\tau(\cdot)\); no per-state quantile shortcut is used.
\medskip

\paragraph{Q-function (temporal-difference target).}
Given the expectile value function $V_\tau$, we train the action–value regressor $\hat Q(s,a)$ by supervised learning on one-step temporal-difference (TD) targets. For each transition $(s_t,a_t,r_t,s_{t+1})$, the TD target is
\[
y_t \;\coloneqq\; r_t \;+\; \gamma\,V_\tau(s_{t+1}),
\]
i.e., immediate reward plus the discounted value of the next state. We then fit $\hat Q$ on concatenated $[s_t,a_t]$ features by minimizing $\sum_t\bigl(\hat Q(s_t,a_t)-y_t\bigr)^2$. Thus, $y_t$ serves as a bootstrapped estimate of $Q^\pi(s_t,a_t)$ derived from $V_\tau$, rather than the true $Q$ itself.
\bigskip

\paragraph{Advantages and IQL weights.}
Advantages are computed as \(A_t=\hat Q(s_t,a_t)-V_\tau(s_t)\), and the IQL weights as \(w_t=\exp(A_t/\beta)\) with standard numerical clipping for stability.
\medskip

\paragraph{Policy learning.}
The policy is trained by weighted regression from state to action, one regressor per action dimension, minimizing \(\sum_t w_t\|\pi_\phi(s_t)-a_t\|_2^2\). Design parameters for the policy models are summarized in Table~\ref{tab:policy_iql_params}.
\bigskip

\begin{table}[htb]
\centering
\caption{Design parameters for the IQL XGBoost policy}
\label{tab:policy_iql_params}
\begin{tabular}{lc}
\hline
Feature vector size & 9 (state only) \\
Model type & XGBRegressor \\
Trees / depth & 1200 / 8 \\
Learning rate & 0.03 \\
$\ell_2$ regularisation & $\lambda = 1$ (default) \\
Diagnostics split & 80/20\,\%\\
Hold-out $R^2$ (u\_I / u\_{Tc}) & 0.8988 / 0.8920 \\
\hline
\end{tabular}
\end{table}

\subsubsection{Simulation results (polymerization)}
\begin{figure}[t]
\centering
\includegraphics[width=\textwidth]{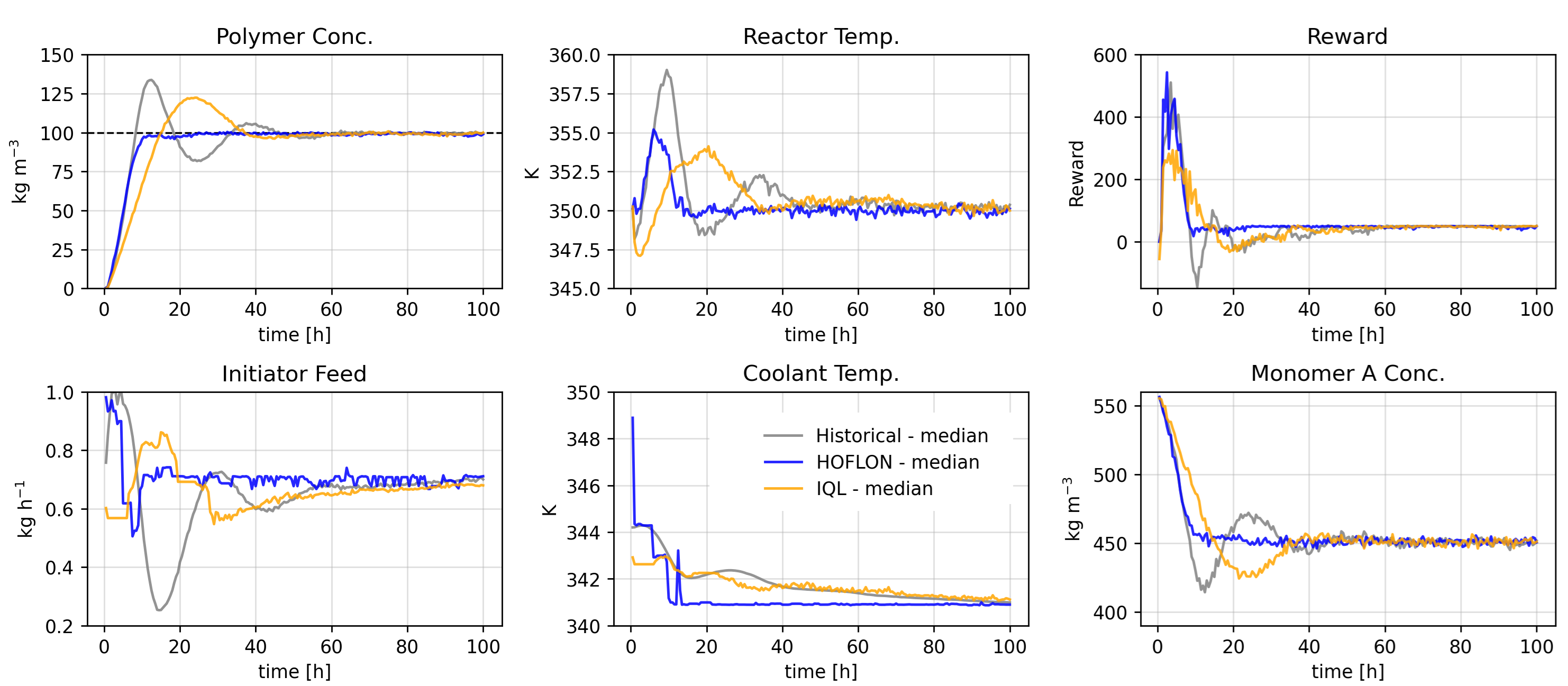}
\caption{Start-up trajectories of the polymerization CSTR under HOFLON, IQL, and the original data generation strategy; shown is the evaluation episode with the median cumulative reward across all rollouts.}
\label{fig:poly_median}
\end{figure}

\begin{figure}[tbh]
\centering
\includegraphics[width=\textwidth]{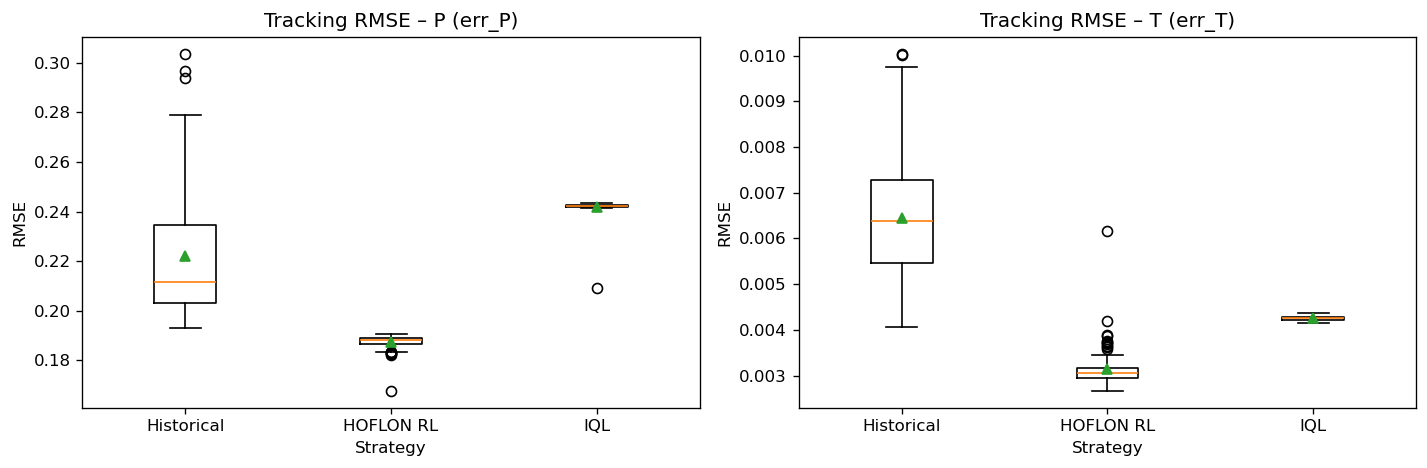}
\caption{Start-up tracking performance comparison of HOFLON, IQL, and the original data generation strategy.}
\label{fig:poly_track_box}
\end{figure}

\begin{figure}[h!]
\centering
\includegraphics[width=0.48\textwidth]{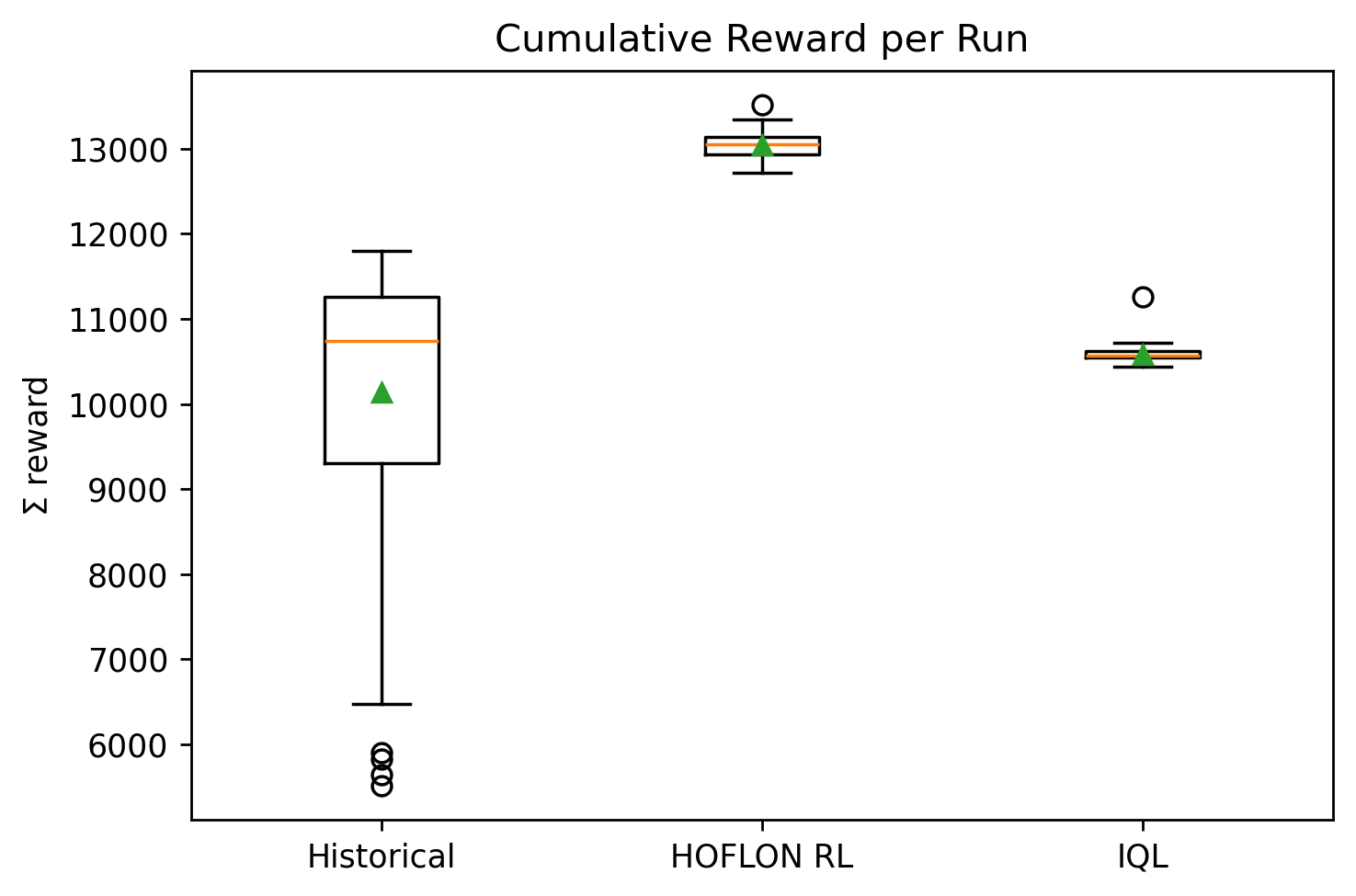}
\caption{Start-up cumulative reward comparison of HOFLON, IQL, and the original data generation strategy}
\label{fig:poly_reward_box}
\end{figure}

We evaluate the proposed HOFLON–RL controller and a trained IQL policy on the polymerization CSTR start-up task. For each controller, we run 100 simulated evaluation episodes under the same operating envelope, initial-condition distribution, and measurement noise used to generate the offline training data. Fig.~\ref{fig:poly_median} presents representative trajectories for the evaluation episode with the median cumulative reward across all rollouts, comparing HOFLON, IQL, and the original data-generation strategy. The distribution of start-up tracking performance across the 100 episodes is summarized in Fig.~\ref{fig:poly_track_box}. Fig.~\ref{fig:poly_reward_box} reports the distribution of cumulative rewards for the same set of episodes. A detailed interpretation of these results is deferred to the Discussion section.


\subsection{Paper machine grade change} 
The evaluation protocol for the paper machine grade change problem mirrors the polymerization CSTR study; here we will only note settings specific to the paper machine. Fig. \ref{fig:train_data_poly} depicts the 100 simulated episodes of the paper machine each driven by a set of PI controllers whose gains were sampled at random.
 
\begin{figure}[t]
    \centering
    \includegraphics[width=\textwidth]{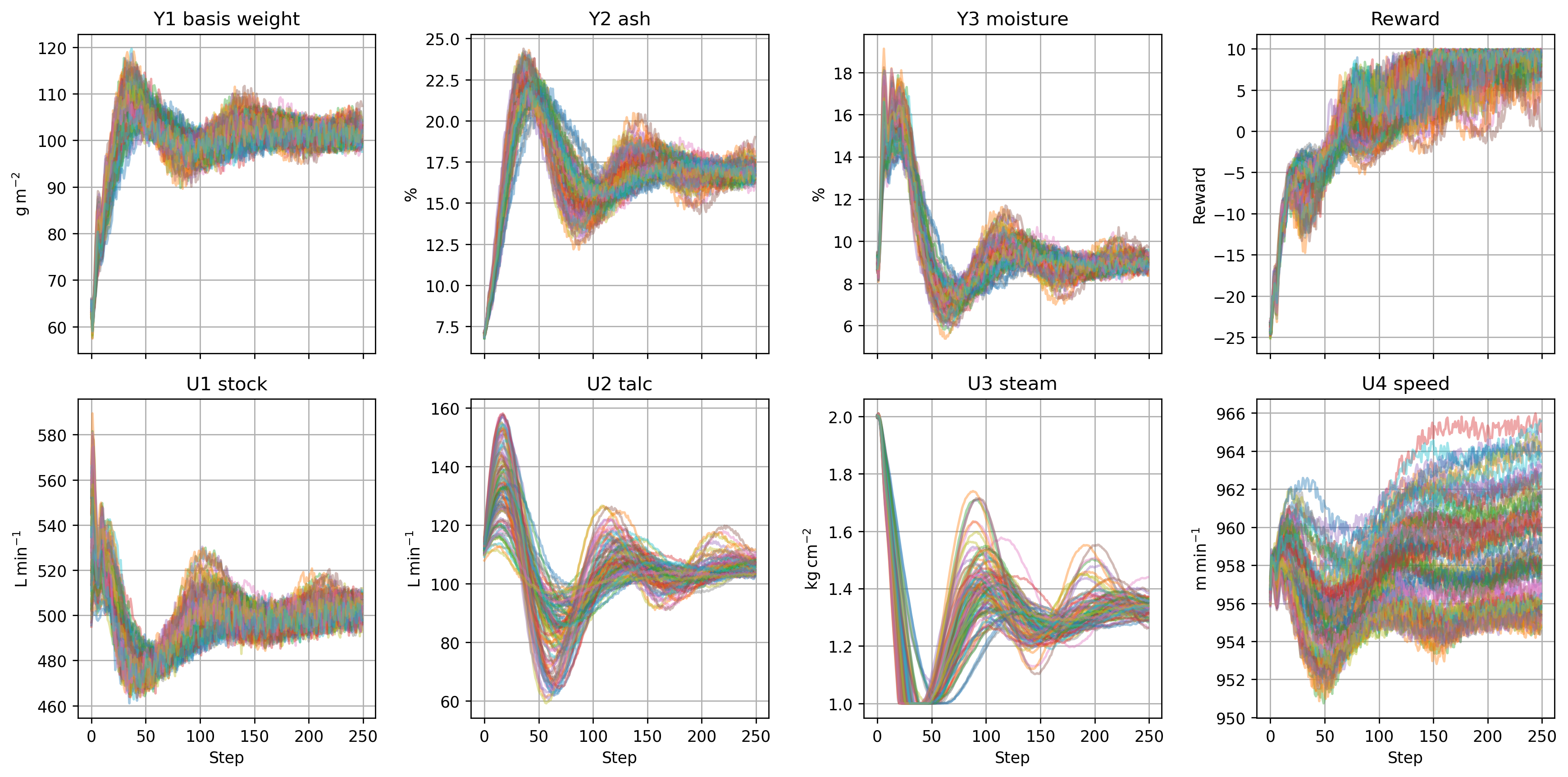}
    \caption{100 PI controlled episodes from the paper machine grade change scenario.}
    \label{fig:train_data_poly}
\end{figure}
 
\subsubsection{HOFLON: offline learning (paper machine)}
\label{sec:offline_stage}

\begin{table}[htb]
\centering
\caption{Design parameters for the XGBoost Q-critic}
\label{tab:q_params_paper}
\begin{tabular}{lc}
\hline
Feature vector size & 14 (10 state + 4 action) \\
Discount factor & $\gamma = 0.90$ \\
Model type & XGBRegressor \\
Trees / depth & 1200 / 12 \\
Learning rate & 0.02 \\
$\ell_2$ regularisation & $\lambda = 1.0$ \\
Train/validation split & 90 / 10\,\% \\
Hold-out $R^2$ & 0.95 \\
\hline
\end{tabular}
\end{table}

\paragraph{Q-critic.}
The action–value function ($Q$-critic) is approximated with an XGBoost regressor trained to predict discounted returns ($\gamma=0.9$). The regressor input is a 14-dimensional feature vector formed by concatenating the ten-element state observation $[Y_1,Y_2,Y_3,e_1,e_2,e_3,\texttt{int1},\texttt{int2},\texttt{int3},\texttt{int4}]$ with the four-element action vector $[U_1,U_2,U_3,U_4]$ (columns in this order). Features are standardized (z-score) using a \texttt{StandardScaler} fitted on the training split before both training and inference. The key hyperparameters for the model, such as the number of trees and learning rate, are summarized in Table~\ref{tab:q_params_paper}. Following training, the model showed very good generalization, achieving a coefficient of determination ($R^2$) of 0.95 on a 20\,\% hold-out validation set. A visualization of the learned $Q$-function for the paper machine case is not provided in this case due to the higher dimension of the problem but we expect the hypersurface to display similar characteristics as in the first case study.
\begin{figure}[t!]
\centering
\includegraphics[width=\textwidth]{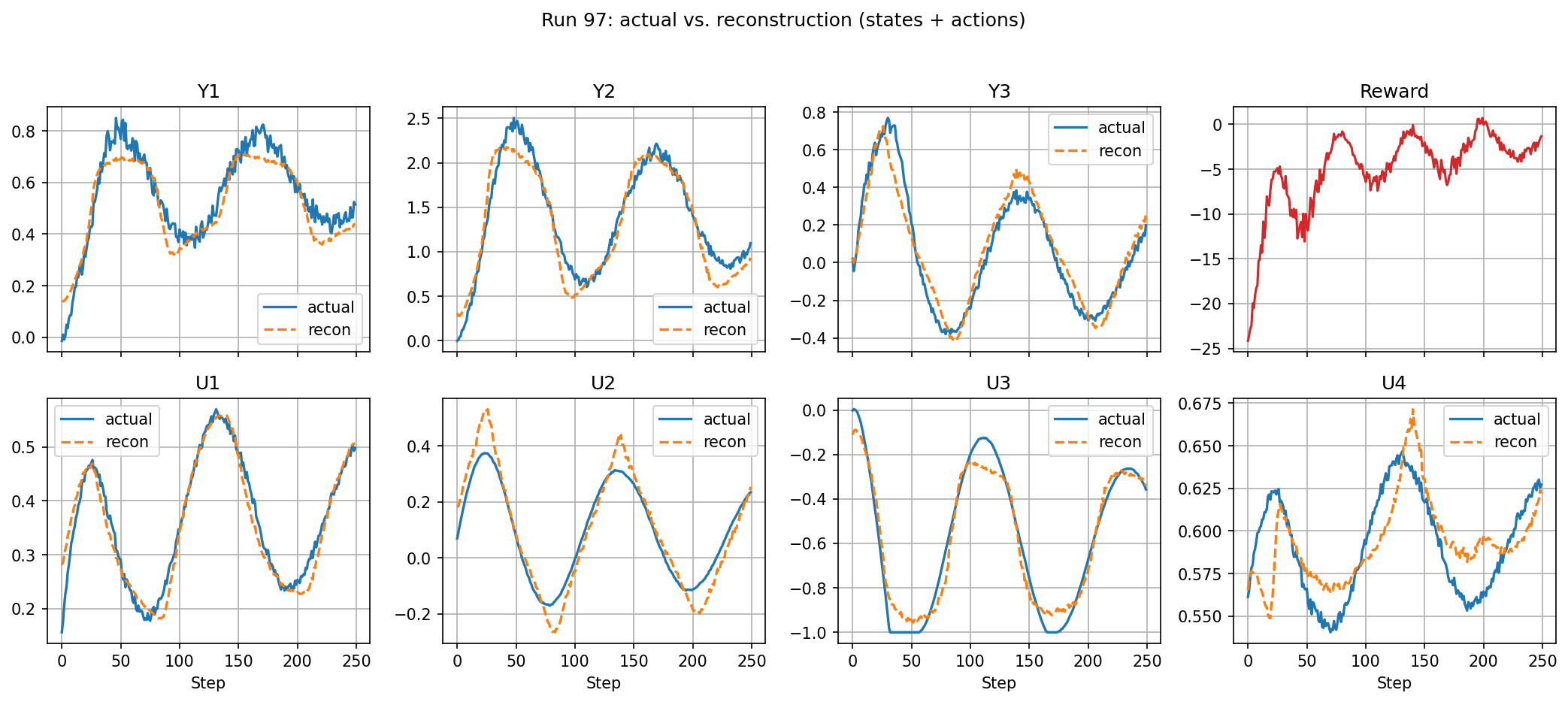}
\caption{Reconstruction of a held-out run using the trained AEE for the controlled variables and the actions.}
\label{fig:aae_recon}
\end{figure}
\begin{table}[h!]
\centering
\caption{Network architecture and training settings for the AAE for the paper machine}
\label{tab:aae_arch}
\begin{tabular}{@{}llcc@{}}
\toprule
\textbf{Component} & \textbf{Layer} & \textbf{Output Dim.} & \textbf{Activation} \\
\midrule
\bfseries Encoder & Input & 14 & -- \\
& FC & 64 & Tanh \\
& FC & 16 & Tanh \\
& FC (latent $z$) & 4 & -- \\
\addlinespace
\bfseries Decoder & FC & 16 & Tanh \\
& FC & 64 & Tanh \\
& FC & 14 & -- \\
\addlinespace
\bfseries Discriminator & FC & 32 & LeakyReLU(0.2) \\
& FC & 16 & LeakyReLU(0.2) \\
& FC & 1 & Sigmoid \\
\midrule[\heavyrulewidth]
\multicolumn{2}{@{}l}{\bfseries Hyperparameter} & \multicolumn{2}{l}{\bfseries Value} \\
\midrule
\multicolumn{2}{@{}l}{Epochs} & \multicolumn{2}{l}{500 (patience 20)} \\
\multicolumn{2}{@{}l}{Test MSE} & \multicolumn{2}{l}{$1.7 \times 10^{-2}$} \\
\multicolumn{4}{@{}c@{}}{\emph{All other entries are identical to the previous case study.}}\\
\bottomrule
\end{tabular}
\end{table}

\paragraph{Adversarial auto-encoder.}
As in the previous case study, we use an AAE to obtain a compact generative manifold; here we only note settings specific to the paper-machine problem. The input is a 14-dimensional vector comprising the CVs $[Y_1,Y_2,Y_3]$, error signals $[e_1,e_2,e_3]$, PI integrator states $[\texttt{int1}\!:\!\texttt{int4}]$, and MVs $[U_1\!:\!U_4]$, scaled to $[-1,1]$ using a \texttt{MinMaxScaler} fit on the training runs only (IDs $0$–$94$). The encoder is a linear MLP $14\!\to\!64\!\to\!16\!\to\!4$ with $\tanh$ activations on hidden layers and a 4-D latent code; the decoder mirrors this as $4\!\to\!16\!\to\!64\!\to\!14$; the discriminator is $4\!\to\!32\!\to\!16\!\to\!1$ with LeakyReLU(0.2) and a sigmoid output. We use Xavier initialisation, MSE reconstruction loss, and a vanilla GAN objective (BCE) for the adversarial term (no gradient penalty), optimized with Adam at $10^{-5}$ for both autoencoder and discriminator, batch size 256, for 500 epochs. Train/test are split by run ID (train: $0$–$94$; test: $95$–$99$); the saved artifacts are the fitted scaler and the encoder/decoder weights.

\subsubsection{HOFLON: online optimization (paper machine)}
\label{sec:online_stage}
\label{sec:online-ho-flon}

\paragraph{Single–step objective.}
As before, at each sampling instant the controller solves the same box-constrained optimization problem given in \ref{eq:online_opt}
The weights $\lambda_1$ and $\lambda_2$ are chosen as 0.45 and 0.0036 respectively.

\paragraph{Solution of the optimization problem.}
\label{sec:solver-portfolio}
As before, we use a derivative-free portfolio, but for the paper-machine case only two local searches are invoked per step, both warm-started from the previous move and with actions clipped to physical bounds $[u_{\min},u_{\max}]\subset\mathbb{R}^4$:
\begin{enumerate}
  \item \textbf{Nelder--Mead simplex} (unconstrained; candidate clipped before $J$ is evaluated).
  \item \textbf{Powell's conjugate-direction} (unconstrained; candidate likewise clipped).
\end{enumerate}

\subsubsection{IQL agent training (paper machine)}
We retain the pipeline from the previous case study, where the Q-function estimation from HOFLON design step was re-used; the policy model settings are summarized in Table~\ref{tab:policy_iql_params_paper}.

\begin{table}[htb]
\centering
\caption{Design parameters for the XGBoost policy (IQL-weighted)}
\label{tab:policy_iql_params_paper}
\begin{tabular}{lc}
\hline
Feature vector size & 10 (state only) \\
Model type & XGBRegressor \\
Trees / depth & 800 / 8 \\
Learning rate & 0.05 \\
$\ell_2$ regularization & $\lambda = 1$ (default) \\
Diagnostics split & Train/hold-out by run ID (last 5 runs held out) \\
Hold-out $R^2$ (U\textsubscript{1}/U\textsubscript{2}/U\textsubscript{3}/U\textsubscript{4}) & 0.8757 / 0.8280 / 0.9383 / 0.8786 \\
\hline
\end{tabular}
\end{table}

\subsubsection{Simulation results (paper machine)}
\begin{figure}[t]
\centering
\includegraphics[width=\textwidth]{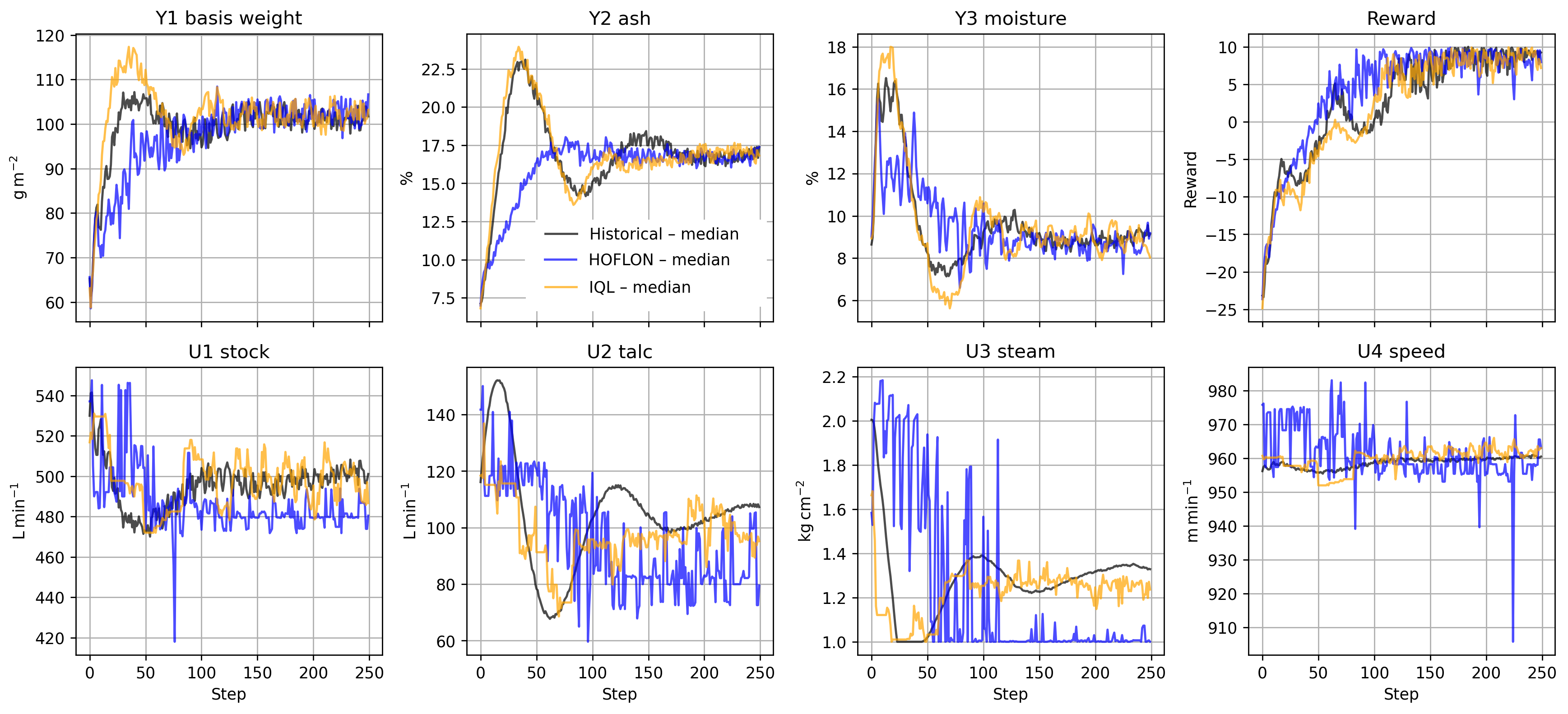}
\caption{Paper grade-change trajectories under HOFLON, IQL, and the original data generation strategy; shown is the evaluation episode with the median cumulative reward across all rollouts.}
\label{fig:paper_median}
\end{figure}

\begin{figure}[tbh]
\centering
\includegraphics[width=\textwidth]{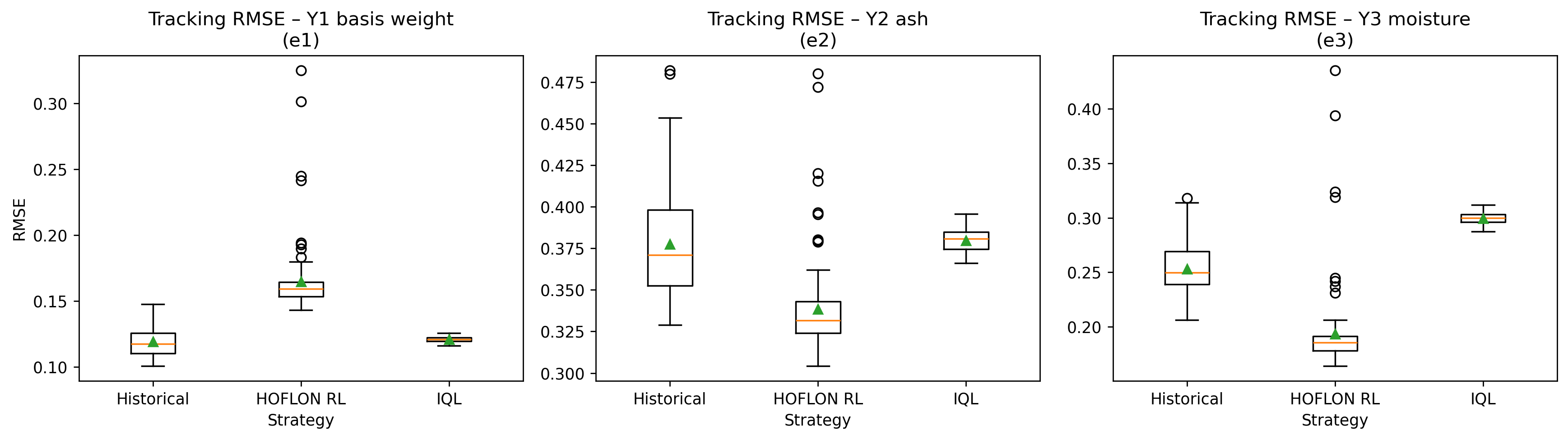}
\caption{Paper grade-change tracking performance comparison of HOFLON, IQL, and the original data generation strategy.}
\label{fig:paper_track_box}
\end{figure}

\begin{figure}[h!]
\centering
\includegraphics[width=0.48\textwidth]{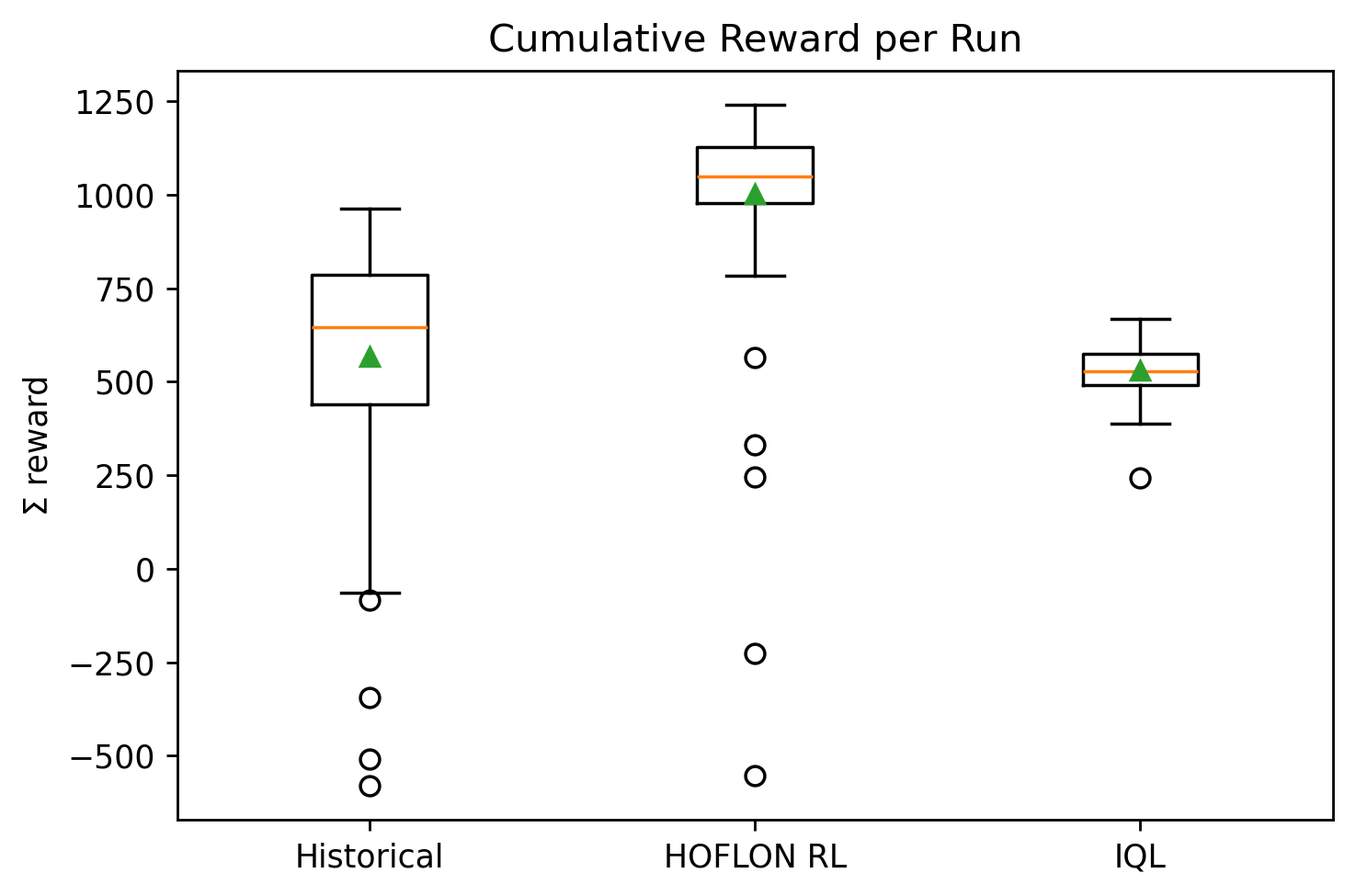}
\caption{Paper grade-change cumulative reward comparison of HOFLON, IQL, and the original data generation strategy}
\label{fig:paper_reward_box}
\end{figure}

We evaluate the proposed HOFLON–RL controller and a trained IQL policy on the paper machine grade chnage problem. For each controller, we run 100 simulated evaluation episodes under the same operating envelope, initial-condition distribution, and measurement noise used to generate the offline training data. Fig.~\ref{fig:paper_median} presents representative trajectories for the evaluation episode with the median cumulative reward across all rollouts, comparing HOFLON, IQL, and the original data-generation strategy. The distribution of start-up tracking performance across the 100 episodes is summarized in Fig.~\ref{fig:paper_track_box}. Fig.~\ref{fig:paper_reward_box} reports the distribution of cumulative rewards for the same set of episodes. A detailed interpretation of these results is deferred to the Discussion section.


\section{Discussion}
\label{sec:Discussion}
Across both benchmarks, HOFLON delivers the highest cumulative reward with low dispersion across episodes, indicating that the decision–time optimization produces consistently good sequences. On the \emph{polymerization start-up}, Table~\ref{tab:cross_case_summary} shows that HOFLON attains the lowest tracking error on both tracked variables ($e_P$ and $e_T$) and the highest cumulative reward among all strategies. This suggests that, for this system, the value critic and the manifold regularization are complementary rather than competing: the critic reliably identifies high-return directions within the portion of the data manifold relevant to start-up, and the move penalty suppresses the impulsive corrections that would otherwise excite thermal dynamics. IQL, in contrast, exhibits higher RMSE for both variables and lower return, consistent with a deterministic policy that hews closely to the behavior manifold and forgoes slightly off-manifold action combinations that the critic judges more valuable; the performance of the historical policy lies between the two in both error and reward.

On the \emph{paper-machine grade change} problem, as summarized in Table~\ref{tab:cross_case_summary}: HOFLON achieves substantially higher return than both the historical controller and IQL while exhibiting an interpretable tracking pattern. Ash (Y\textsubscript{2}) and moisture (Y\textsubscript{3}) RMSEs are reduced relative to both comparators, whereas basis-weight (Y\textsubscript{1}) RMSE increases modestly. This trade-off aligns with the objective exposed at action selection: the value term pulls toward trajectories that the critic deems high-return; the data-manifold term (AAE) discourages excursions into poorly supported regions of state–action space; and the move penalty tempers aggressive adjustments that could destabilize the process or over-travel actuators. Given the SIMO pairing (Y\textsubscript{1} driven by both $U_1$ and $U_4$), these ingredients can prioritize moisture and ash regulation when sharp basis-weight corrections would erode overall return—mirroring plant economics where transient moisture/ash excursions are often penalized more heavily than brief, small deviations in basis weight. It is also worth noting that, unlike the polymerization case—which exposes internal states to the learner—the paper-machine study trains only on measured outputs, errors, and integrator values, and operates in a larger MIMO setting (four inputs, three outputs). The lack of state measurements and the higher dimensionality make the critic harder to fit and the online objective rougher to optimize, which likely explains the overall weaker tracking levels observed on this task despite the clear reward advantage.

\begin{table}[t!]
\centering
\caption{Summary across both case studies (mean $\pm$ std over 100 episodes)}
\label{tab:cross_case_summary}
\begin{tabular}{lcccc}
\toprule
\multicolumn{5}{c}{\textbf{Polymerization start-up}}\\
\midrule
\textbf{Strategy} & \textbf{RMSE $e_P$} & \textbf{RMSE $e_T$} & \textbf{Cumulative reward} \\
\midrule
Historical & $0.222 \pm 0.026$ & $0.006 \pm 0.001$ & $10135 \pm 1561$ \\
IQL        & $0.242 \pm 0.003$ & $0.004 \pm 0.000$ & $10583 \pm 86$ \\
HOFLON RL  & $\mathbf{0.187 \pm 0.003}$ & $\mathbf{0.003 \pm 0.000}$ & $\mathbf{13044 \pm 139}$ \\
\midrule
\multicolumn{5}{c}{\textbf{Paper-machine grade change}}\\
\midrule
\textbf{Strategy} & \textbf{RMSE Y\textsubscript{1}} & \textbf{RMSE Y\textsubscript{2}} & \textbf{RMSE Y\textsubscript{3}} & \textbf{Cumulative reward} \\
\midrule
Historical & $\mathbf{0.119 \pm 0.011}$ & $0.378 \pm 0.034$ & $0.253 \pm 0.024$ & $10135 \pm 1561$ \\
IQL        & $0.121 \pm 0.002$ & $0.380 \pm 0.007$ & $0.300 \pm 0.005$ & $10583 \pm 86$ \\
HOFLON RL  & $0.165 \pm 0.026$ & $\mathbf{0.338 \pm 0.029}$ & $\mathbf{0.194 \pm 0.039}$ & $\mathbf{13044 \pm 139}$ \\
\bottomrule
\end{tabular}
\end{table}

\paragraph{Methodological interpretation}
Conceptually, HOFLON performs a \emph{regularized greedy policy-improvement step} at each decision point. Pure greed—maximizing a learned critic—can be brittle in offline RL because function approximators may overestimate value in out-of-distribution regions. The AAE manifold therefore acts as a behavior regularizer: it keeps the optimizer near the logged data support unless the critic indicates sufficiently large value gains to justify a controlled deviation. The move penalty adds an orthogonal regularizer that smooths control actions and suppresses myopic jumps. This combination explains why HOFLON realizes higher returns than IQL without sacrificing stability: the method is “greedy where safe, conservative where uncertain.” By comparison, IQL trains a deterministic policy via weighted regression toward dataset actions, guided by an expectile value function that stays in-distribution; this yields low-variance behavior but tends to forgo slightly off-manifold action combinations that the critic judges more valuable, resulting in narrower dispersion but systematically lower cumulative reward.

\paragraph{Computational feasibility}
Warm-started, derivative-free searches remain lightweight for the action dimensions considered, keeping wall-clock time comfortably below the sampling periods. Although local search on a nonconvex objective can be sensitive to initialization, warm starts from the previous move provide continuity, while the manifold term effectively smooths the search landscape by discouraging large excursions into poorly supported regions. In the polymerization case, adding a light global scan further reduced the chance of poor local optima; the same mechanism can be enabled for the paper machine if future tuning indicates benefit.

\paragraph{Weighting the objective: practical tuning challenges}
A practical challenge is choosing the relative weights between value maximization, manifold consistency, and move suppression. Even with normalized features and actions, these terms have different scales and interact nonlinearly through the plant dynamics: increasing the manifold penalty can hide local critic overestimation but may also discourage beneficial exploratory moves; strengthening the move penalty improves smoothness and actuator hygiene but can slow convergence; and amplifying the value term can raise return while making the optimizer more sensitive to local roughness in $\hat Q$. Here we selected a single set of weights per case study and did not perform fine-grained, channel-wise tuning; the reported performance is therefore a conservative lower bound. Automated schedules or adaptive penalty rules are promising next steps.

\paragraph{Model choices under data and runtime constraints}
Model choice was deliberately simplified. We adopted XGBoost for the critic and for the IQL policy regressor as a pragmatic bias–variance compromise that trains quickly, offers strong tabular performance, and keeps inference cost modest during online optimization. We experimented with Gaussian-process regression using a 500-point subset (out of $\sim 25{,}000$ samples) to keep cubic training cost manageable; the GP produced a smooth surface that was easy to optimize online, but held-out $R^2$ was poor and closed-loop returns degraded. Lightweight fully connected neural networks underfit and did not perform well; we chose not to scale to larger networks given added training complexity and the risk of sharper nonconvexities in the decision-time objective. Nonetheless, the piecewise-constant nature of boosted trees induces a non-smooth objective; in principle, replacing the ensemble with a smoother regressor that preserves predictive accuracy (e.g., splines, calibrated shallow NNs, or sparse/inducing-point GPs) could ease optimization and improve returns. We view this as a promising avenue for future work.

\paragraph{Per-channel weighting and reward design}
We did not tune individual term weights per controlled variable. In the paper-machine problem, for example, a CV-specific move penalty or manifold weight could temper the observed trade-off on basis weight without compromising ash and moisture regulation. Such per-channel weighting—and schedules that adapt penalties as the controller approaches target—are natural extensions. Reward design is similarly central. We used a dual \emph{progress-and-proximity} reward: one component rewards directional progress toward the target; the other rewards staying near the target once reached. This structure balanced swift transitions with steady regulation and enabled HOFLON to prioritize moves that both approach and hold the set-point corridor. We did not further tune the relative scaling or shaping; in deployment, start-up or grade-change procedures may warrant finer reward definitions (e.g., asymmetric costs, CV-specific corridors, or time-varying priorities) to encode plant economics and safety envelopes more precisely.

\paragraph{Limitations}
Both critic and manifold are trained on finite logs and inevitably carry approximation error; the manifold reduces but does not eliminate the risk that the optimizer exploits local artifacts. The observed increase in Y\textsubscript{1} RMSE on the paper machine is an intentional outcome of the chosen weights and pairing; depending on economics, this balance can be shifted by reweighting the objective or by introducing CV-specific set-point corridors. Results are simulation-based; deployment will require state estimation, actuator rate/travel limits, and online anomaly detection, as well as automated schedules for the value/manifold/move weights to reduce manual tuning effort.

\paragraph{Toward a generalized, transition-agnostic HOFLON}
The results suggest a path toward an operations-wide agent that can handle arbitrary start-ups and grade changes without retraining. A generalized critic can embed the desired transition directly into its input—set-point paths, ramp rates, and quality envelopes—so a single agent plans many transitions. To preserve fast decision-time optimization, convex-in-action architectures such as input-convex neural networks (ICNNs) are attractive: nonlinear in state yet convex in action, reducing runtime search to a single convex program. The representation cost of convexity should be weighed against alternatives (monotone networks, quadratic forms, Gaussian-process hybrids) to find the best accuracy–solve-time trade-off. Conditioning should also be explicit in the generative component: our current AAE imposes a global manifold over $(s,a)$, whereas conditional latent-variable models (e.g., CVAEs) can model the \emph{conditional} action distribution given state and transition intent—potentially improving optimizer conditioning. Finally, testing without plant models or simulators motivates a dual track: offline back-testing on archived transitions to quantify data sufficiency, coverage gaps, and sensor corruption sensitivity; and live shadow trials that log the agent’s proposed moves in parallel with the existing control system. Metrics such as action mismatch, predicted-versus-realized reward, and distance to safety envelopes can be monitored to decide whether recommendations are acceptably close, systematically biased, or occasionally unsafe; robust thresholds and statistical tests for “far-off” suggestions remain an open research problem.

\section{Conclusions}
\label{sec:Conclusions}
We presented HOFLON, a hybrid framework that couples an offline-learned value critic and a data manifold with online, solver-based action selection.\,On two industrially motivated benchmarks—a polymerization start-up and a paper-machine grade change—HOFLON consistently achieved the highest cumulative reward among all policies tested and, in the paper machine, substantially improved ash and moisture tracking at the cost of a modest increase in basis-weight RMSE.\,These results underscore three advantages of the approach: (i) explicit, tuneable trade-offs among performance, safety, and move smoothness at decision time; (ii) reduced OOD risk through manifold regularization; and (iii) real-time feasibility with simple derivative-free searches.\,The framework is broadly applicable to transition operations where accurate first-principles models are unavailable but historical logs exist. Future work will target automatic weight scheduling; compare smoothed/ensembled critics; refine reward shaping; develop OOD and safety diagnostics using the latent penalty; and pursue the development of transition-agnostic critics and conditional manifolds (e.g., ICNNs, CVAEs) to enhance consistent and fast decision-time optimization across start-ups and grade changes.
 
\bibliographystyle{plainnat}
\bibliography{references}
 
\clearpage

\end{document}